\documentclass{article}

\PassOptionsToPackage{numbers, sort, compress}{natbib}

\usepackage[preprint]{neurips_2026}

\usepackage[utf8]{inputenc} % allow utf-8 input
\usepackage[T1]{fontenc}    % use 8-bit T1 fonts
\usepackage{hyperref}       % hyperlinks
\usepackage{url}            % simple URL typesetting
\usepackage{booktabs}       % professional-quality tables
\usepackage{amsfonts}       % blackboard math symbols
\usepackage{nicefrac}       % compact symbols for 1/2, etc.
\usepackage{microtype}      % microtypography
\usepackage{xcolor}         % colors

\usepackage{amsmath}
\usepackage{amssymb}
\usepackage{graphicx}
\usepackage{multirow}
\usepackage{bm}
\usepackage{array}
\usepackage{makecell}
\usepackage{adjustbox}
\usepackage[table]{xcolor}
\usepackage{pifont}
\usepackage{algorithm}
\usepackage{algpseudocode}
\usepackage{wrapfig}
\usepackage{caption}

\title{GeoCore-9B: Towards Geo-Aware Generative Foundation Models in Earth Observation}

\author{
  Jeonghyeok~Do\\
  Information \& Electronics Research Institute\\
  KAIST\\
  \texttt{ehwjdgur0913@kaist.ac.kr}\\
  \And
  Munchurl~Kim\thanks{Corresponding author.}\\
  School of Electrical Engineering\\
  KAIST\\
  \texttt{mkimee@kaist.ac.kr} \\
}

\begin{document}

\maketitle

\vspace{-2.0em}
\begin{center}
    \small
    Project page: \url{https://kaist-viclab.github.io/GeoCore-9B_site/}
\end{center}
\vspace{0.5em}

\begin{abstract}
Existing generative models for earth observation (EO) predominantly rely on fine-tuning natural image priors, which limits their scalability and introduces perspective biases that conflict with geospatial constraints. To address this, we introduce \textbf{GeoCore-9B}, a 9-billion-parameter generative foundation model, which is the first of its scale to be trained from scratch exclusively on EO data. Unlike previous EO foundation models, GeoCore-9B is built upon a Flow Matching-based Diffusion Transformer (DiT) and natively conditions generation on text descriptions and continuous geospatial metadata, including ground sample distances, latitudes, and longitudes. To overcome the convergence and spatial disorientation challenges of training at this scale, we propose a Geospatial Semantic Alignment loss. This objective distills structural Earth surface priors (e.g., terrain and urban areas) from a frozen specialist teacher network, constraining the diffusion latent trajectory during training without adding inference overhead. Pre-trained on the global-scale Git-10M dataset, GeoCore-9B demonstrates strong downstream versatility. Beyond standard proxy generative tasks, we show that GeoCore-9B can be effectively adapted for practical EO applications, including highly challenging tasks such as cloud removal and SAR-to-optical cross-modal translation. Extensive evaluations confirm that GeoCore-9B establishes new state-of-the-art performance in both visual fidelity and geographic structural accuracy.
\end{abstract}

\section{Introduction}

Generative foundation models for Earth Observation (EO) \cite{khanna2024diffusionsat, sebaq2024rsdiff, liu2024diffusion, tang2024crs, liu2025text2earth, yu2024metaearth, jakubik2025terramind} have emerged as a pivotal technology for simulating terrestrial environments, augmenting scarce datasets, and enabling complex downstream applications. Beyond generic image synthesis, a practically useful EO foundation prior should be transferable to restoration and cross-modal translation scenarios, where the models must recover geographically faithful structures from real occlusions, degradations, or sensor-induced modality gaps. However, generating satellite imagery faces unique challenges: unlike natural images, EO data is strictly orthographic, physically anchored by spatial resolution (ground sample distance, GSD), and covers highly dense and heterogeneous geographic structures across the globe. Therefore, establishing a robust generative world model that inherently comprehends these physical and geometric constraints remains an ongoing challenge.

Existing generative approaches \cite{khanna2024diffusionsat, sebaq2024rsdiff, tang2024crs, liu2025text2earth} in the EO domain have primarily relied on \textit{fine-tuning U-Net-based pre-trained models} (e.g., Stable Diffusion v1.5 and v2.1 \cite{rombach2022high}) \textit{originally optimized for natural images}. While this fine-tuning paradigm accelerates convergence, it inevitably introduces severe domain shifts. Natural image priors are inherently biased toward perspective projection, center-object framing, and casual spatial scales, which fundamentally conflict with the scale-invariant, bird's-eye view nature of satellite imagery. Furthermore, the reliance on standard U-Net architectures severely bottlenecks their representational capacity. Consequently, these models often struggle with geometric distortions and fail to capture authentic geospatial data distributions. Another limitation lies in how downstream capability has been evaluated. Existing evaluations often focus on controllability-oriented or proxy generative settings, such as sketch-conditioned EO generation \cite{tang2024crs} and multimodal generation \cite{liu2025text2earth} built by synthetically augmenting the RSICD \cite{lu2017exploring} text-to-image benchmark. While useful for assessing conditional generation, such protocols provide limited evidence that the learned generative prior can transfer to practical EO tasks involving real paired observations, occlusions, degradations, or severe cross-sensor modality gaps. Even a recent attempt \cite{jakubik2025terramind} to build an EO model from scratch remains constrained: its representational capacity is often diluted across broad multimodal tasks, and it is a relatively small-scale model, lacking the massive parameter scale required to synthesize the immense complexity of the Earth's surface.

To address these limitations, we propose \textbf{GeoCore-9B}, a 9-billion-parameter generative foundation model trained from scratch. GeoCore-9B is the first generative foundation model built upon a Flow Matching-based Diffusion Transformer (DiT) \cite{peebles2023scalable, labs2025flux, esser2024scaling} for EO. It is pre-trained on the global-scale Git-10M \cite{liu2025text2earth} dataset and conditions generation on text descriptions and geospatial metadata, including GSD, latitude, and longitude. This design avoids reliance on natural image priors and enables geo-aware EO synthesis. Training such a large model from scratch, however, poses severe convergence and spatial disorientation challenges. To address this, we introduce a \textbf{Geospatial Semantic Alignment} loss, a training-only alignment objective that distills satellite-specific structural cues from a frozen DINOv3-Sat \cite{simeoni2025dinov3} as a teacher network. By aligning intermediate DiT representations with geospatial semantic features, GeoCore-9B improves structural fidelity in generated EO images without adding inference overhead. Beyond pre-training, we validate the transferability of the learned EO foundation prior on practical downstream applications. For example, even with parameter-efficient fine-tuning (e.g., LoRA \cite{hu2022lora}), GeoCore-9B can be adapted to highly challenging tasks such as cloud removal and SAR-to-optical translation, demonstrating its efficacy and utility beyond controllability-oriented or synthetic proxy generation tasks. The main contributions of our work are summarized as follows:

\begin{itemize}
    \item We introduce \textbf{GeoCore-9B}, a 9-billion-parameter DiT-based generative foundation model trained from scratch on EO data with text and geospatial metadata.
    \item We propose a \textbf{Geospatial Semantic Alignment} loss, which uses a frozen satellite-specialist teacher to improve structural fidelity during training with zero inference overhead.
    \item We demonstrate practical downstream transferability by adapting GeoCore-9B to cloud removal and SAR-to-optical translation, where it outperforms or remains competitive compared to task-specific specialist methods.
\end{itemize}

\section{Related Work}

\textbf{Generative models in Earth Observation (EO).} Generative modeling for EO \cite{khanna2024diffusionsat, sebaq2024rsdiff, liu2024diffusion, tang2024crs, liu2025text2earth, yu2024metaearth, jakubik2025terramind} has evolved from adopting natural-image diffusion models to developing domain-specific and multimodal frameworks. 
DiffusionSat \cite{khanna2024diffusionsat} adapts U-Net-based pre-trained models \cite{rombach2022high} by incorporating temporal and multi-spectral conditions for satellite image generation. 
RS-Diff \cite{sebaq2024rsdiff} proposes a cascaded architecture that sequentially generates and super-resolves remote sensing imagery. 
CRS-Diff \cite{tang2024crs} improves controllability by injecting composite spatial signals, such as sketches and semantic masks, through multi-scale feature fusion. 
Text2Earth \cite{liu2025text2earth} scales text-driven EO generation with the global-scale Git-10M dataset, while TerraMind \cite{jakubik2025terramind} introduces an any-to-any multimodal framework with a unified transformer backbone. 
Our proposed GeoCore-9B further scales generative pre-training from scratch on EO data and incorporates geospatial metadata for geo-aware synthesis.

\textbf{Scalable generative architectures and alignment.} Large-scale image generation has shifted from U-Net-based diffusion models~\cite{rombach2022high} to Diffusion Transformers (DiT)~\cite{peebles2023scalable}, as demonstrated by recent Flow Matching-based models such as Stable Diffusion v3.0~\cite{esser2024scaling} and FLUX~\cite{labs2025flux}. 
Flow Matching~\cite{liu2022flow, lipman2022flow} learns a continuous vector field from noise to data, offering a scalable and stable training objective for large generative models. 
In parallel, representation alignment has been shown to improve diffusion training by aligning generative features with strong visual representations~\cite{yu2024representation, leng2025repa, yao2025reconstruction}. 
We build our GeoCore-9B on these advances by combining a Flow Matching-based DiT backbone with a training-only geospatial alignment objective specifically tailored to satellite imagery.

\section{GeoCore-9B}

\begin{figure}[t]
  \centering
  \includegraphics[width=\textwidth]{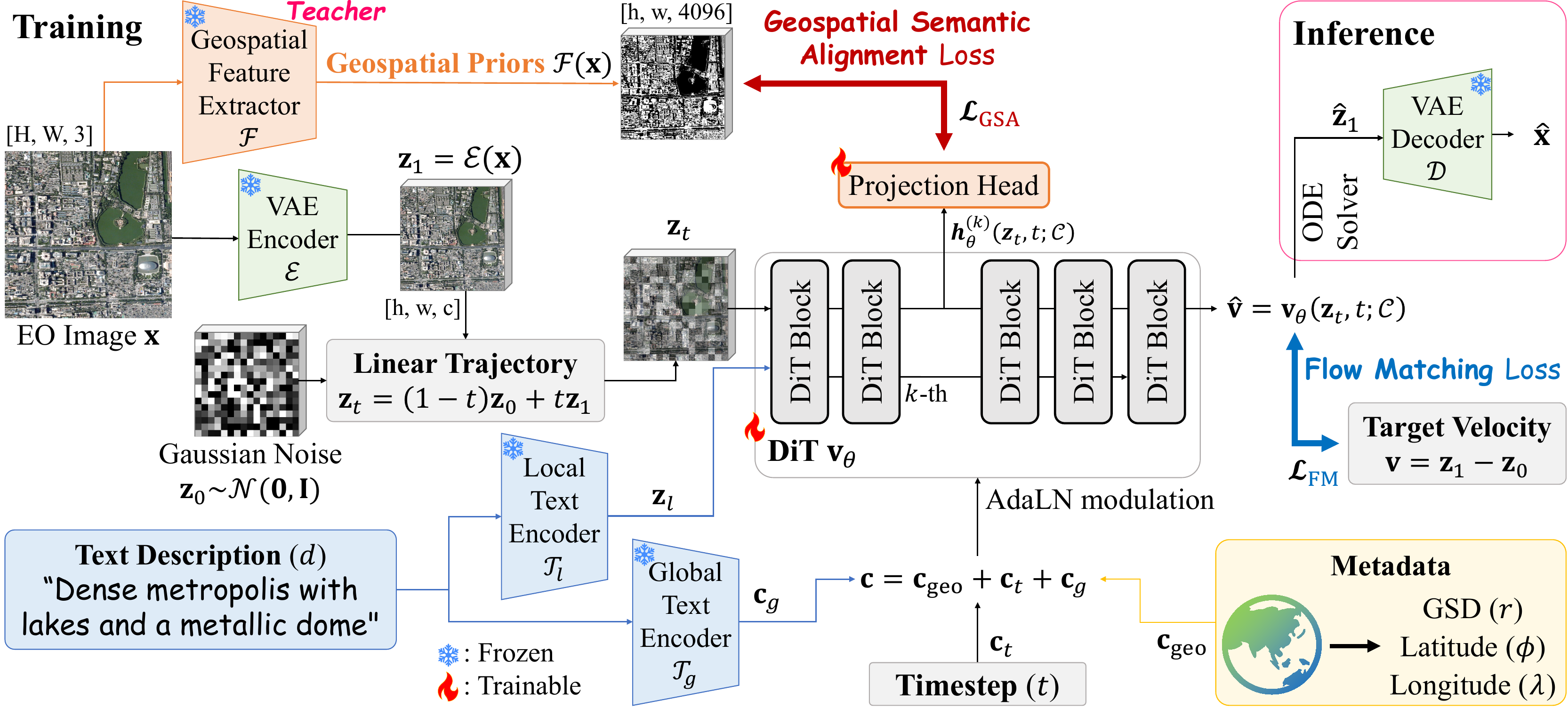}
  \caption{
  Overview of GeoCore-9B.
  GeoCore-9B trains a Flow Matching-based DiT in the latent space, conditioned on local text tokens and global geospatial metadata through token fusion and AdaLN.
  A training-only Geospatial Semantic Alignment loss distills structural Earth priors from a frozen satellite-specialist teacher, improving spatial fidelity without adding inference overhead.
  }
  \label{fig:framework}
\end{figure}

Fig.~\ref{fig:framework} illustrates the conceptual flow of our GeoCore-9B built upon a Flow Matching-based DiT~\cite{peebles2023scalable, labs2025flux, esser2024scaling}, augmented with text and geospatial metadata conditioning, alongside a training-only semantic alignment objective.

\subsection{Geo-Conditioned Flow Matching Backbone}

GeoCore-9B operates in the latent space of a pre-trained VAE~\cite{labs2025flux}.
Given an RGB image $\mathbf{x} \in \mathbb{R}^{H \times W \times 3}$, we obtain its latent representation $\mathbf{z}_1=\mathcal{E}(\mathbf{x})$.
Following Flow Matching \cite{liu2022flow, lipman2022flow}, we sample $\mathbf{z}_0 \sim \mathcal{N}(\mathbf{0},\mathbf{I})$ and define a linear trajectory:
\begin{equation}
    \mathbf{z}_t = (1-t)\mathbf{z}_0 + t\mathbf{z}_1, \quad t \in [0,1],
\end{equation}
where the target velocity is $\mathbf{v}=\mathbf{z}_1-\mathbf{z}_0$.
The DiT backbone $\mathbf{v}_{\theta}$ is trained to predict $\mathbf{v}$ from the noisy latent $\mathbf{z}_t$ under text and geospatial conditions. To condition generation on a text description $d$, we use a global text embedding $\mathcal{T}_g(d)$ from CLIP-ViT-L/14~\cite{radford2021learning} and a token-level text embedding $\mathcal{T}_l(d)$ from T5-XXL~\cite{raffel2020exploring} as:
\begin{equation}
    \mathbf{c}_g = \mathcal{T}_g(d) \in \mathbb{R}^{1 \times c_g},
    \qquad
    \mathbf{z}_l = \mathcal{T}_l(d) \in \mathbb{R}^{M \times c_l},
\end{equation}
where $M$ is the text sequence length.
After linear projection, $\mathbf{z}_l$ is concatenated with the image latent tokens $\mathbf{z}_t$ and processed by the DiT blocks, while $\mathbf{c}_g$ is used for global modulation. We use 3D rotary positional embeddings (RoPE) \cite{su2024roformer} to encode both text-token positions and latent spatial coordinates.

\subsection{Geospatial Metadata Conditioning}
\label{sec:geo_metadata}

Satellite RGB imagery is strongly tied to physical scale and geographic location.
Given a GSD $r$, latitude $\phi$, and longitude $\lambda$, we map each scalar with a sinusoidal projection $\Phi(\cdot)$ and combine them to obtain the geospatial context vector $\mathbf{c}_{\mathrm{geo}}$ as:
\begin{equation}
    \mathbf{c}_{\mathrm{geo}}
    =
    \mathrm{MLP}_{r}\left(\Phi(r)\right) + \mathrm{MLP}_{\phi}\left(\Phi(\phi)\right) + \mathrm{MLP}_{\lambda}\left(\Phi(\lambda)\right) .
\end{equation}
Then, we obtain a global conditioning vector $\mathbf{c}$ by adding $\mathbf{c}_{\mathrm{geo}}$, the timestep embedding $\mathbf{c}_t$, and the global text embedding $\mathbf{c}_g$ as $\mathbf{c} = \mathbf{c}_{\mathrm{geo}} + \mathbf{c}_t + \mathbf{c}_g$, where $\mathbf{c}$ modulates the DiT activations through AdaLN~\cite{peebles2023scalable}.
For classifier-free guidance~\cite{ho2022classifier}, we apply condition dropout to the text and geospatial metadata, using learnable null embeddings for missing metadata conditions.

\subsection{Geospatial Semantic Alignment}
\label{sec:gsa}

Training a 9B-parameter DiT from scratch on EO data is challenging due to slow convergence and spatially unstable generation.
To stabilize training, we introduce a Geospatial Semantic Alignment (GSA) loss, a training-only feature alignment objective.
We use a frozen DINOv3-Sat \cite{simeoni2025dinov3} encoder $\mathcal{F}(\cdot)$ as a satellite-specialist teacher and align intermediate DiT features with its dense structural representations.
Given intermediate latent tokens $\mathbf{h}_{\theta}^{(k)}(\mathbf{z}_t,t,\mathcal{C})$ at layer $k$, the GSA loss is defined as:
\begin{equation}
    \mathcal{L}_{\mathrm{GSA}}
    =
    \mathbb{E}_{\mathbf{z}_0,\mathbf{x},t,\mathcal{C}}
    \left[
    \left\|
    \mathbf{W}_{\mathrm{proj}}
    \left(
    \mathbf{h}_{\theta}^{(k)}(\mathbf{z}_t,t,\mathcal{C})
    \right)
    -
    \mathcal{F}(\mathbf{x})
    \right\|_2^2
    \right],
\end{equation}
where $\mathcal{C} = \{d, r, \phi, \lambda\}$ is the conditioning set and $\mathbf{W}_{\mathrm{proj}}$ maps the DiT features to the teacher feature dimension.
Since $\mathcal{F}$ and $\mathbf{W}_{\mathrm{proj}}$ are used only during training, the GSA loss improves structural fidelity without increasing inference cost.

\subsection{Training Objective}

The final training objective combines the Flow Matching loss ($\mathcal{L}_{\mathrm{FM}}$) and the GSA loss as:
\begin{equation}
    \mathcal{L}_{\mathrm{total}}
    =
    \underbrace{
    \mathbb{E}_{\mathbf{z}_0,\mathbf{z}_1,t,\mathcal{C}}
    \left[
    \left\|
    \mathbf{v}_{\theta}(\mathbf{z}_t,t,\mathcal{C})
    -
    (\mathbf{z}_1-\mathbf{z}_0)
    \right\|_2^2
    \right]
    }_{\mathcal{L}_{\mathrm{FM}}}
    +
    \mu \mathcal{L}_{\mathrm{GSA}},
\end{equation}
where $\mu$ controls the power of the semantic alignment and is empirically set to $0.5$ in our experiments.

\begin{figure}[t]
  \centering
  \includegraphics[width=1.0\textwidth]{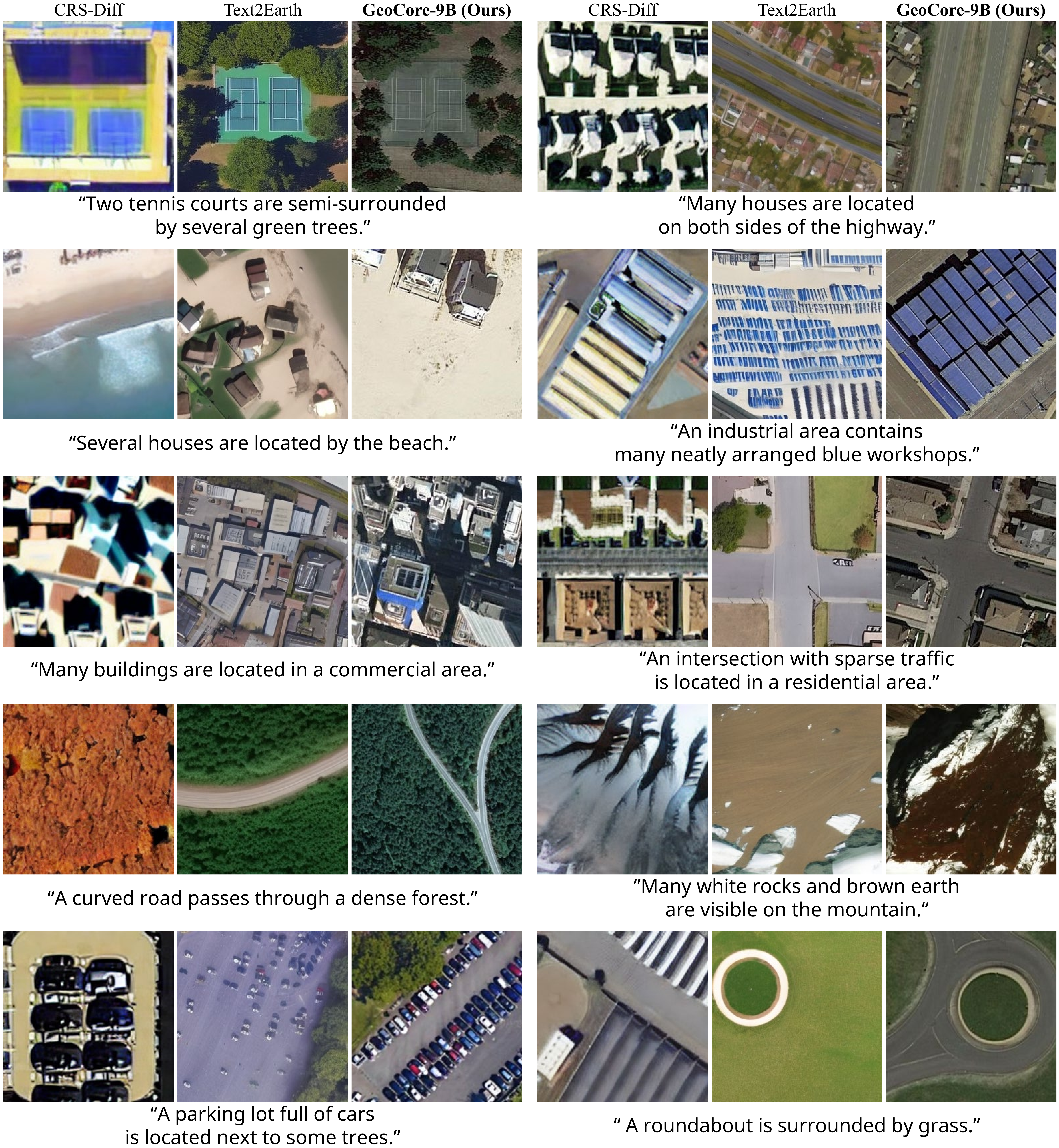}
\caption{
    Qualitative comparison of text-conditioned generation. Given various text descriptions, GeoCore-9B synthesizes highly realistic and structurally accurate satellite imagery, outperforming baselines (CRS-Diff \cite{tang2024crs} and Text2Earth \cite{liu2025text2earth}) which often suffer from severe artifacts.
    }
    \label{fig:text_cond}
\end{figure}

\section{Experiments}

\subsection{Datasets}

We evaluate GeoCore-9B on both generative and practical downstream EO tasks.
For pre-training, we use Git-10M~\cite{liu2025text2earth}, a global-scale satellite RGB image dataset containing 10M image-text pairs with geospatial metadata, including GSDs, latitudes, and longitudes.
For text-to-image adaptation, we use RSICD~\cite{lu2017exploring}, which contains 10,921 remote sensing image-text pairs across 30 scene categories but does not provide geospatial metadata.
For practical downstream evaluation, we use Sen2-MTC~\cite{huang2022ctgan} for cloud removal and QXS-SAROPT~\cite{huang2021qxs} for SAR-to-optical image translation.

\subsection{Implementation Details}

\textbf{Pre-training.}
Input images are randomly cropped to $256 \times 256 \times 3$ and encoded into a $32 \times 32 \times 32$ latent space using a pre-trained VAE encoder~\cite{labs2025flux}.
After $2 \times 2$ patchification, the $16 \times 16 \times 128$ latent token grid is projected to a hidden dimension of $4,096$.
GeoCore-9B uses $32$ DiT blocks with $32$ attention heads, an MLP ratio of $3.0$, and 3D RoPE~\cite{su2024roformer} axis dimensions of 32 (text), 48 (image height), and 48 (image width).
Token-level text embeddings are truncated or padded to a maximum sequence length of $M=256$ and have a dimension of $c_l=4,096$, while the global text embedding has a dimension of $c_g=768$; both are projected to the DiT hidden dimension before conditioning.
The timestep, GSD, longitude, and latitude conditions are each encoded as $256$-dimensional sinusoidal features, and are projected to the hidden dimension through separate MLP embedders.
For the GSA loss, the frozen DINOv3-Sat~\cite{simeoni2025dinov3} teacher produces $16 \times 16 \times 4,096$ dense features, and we apply the alignment loss at the $k=8$-th DiT block after projecting the intermediate DiT features to the teacher feature space, with $\mu=0.5$.
We train GeoCore-9B from scratch on the full Git-10M dataset for 300K iterations using AdamW with a constant learning rate of $1\times10^{-4}$, a weight decay of $0.001$, a global batch size of $1,024$, and bfloat16 (bf16) mixed precision.
Following Text2Earth~\cite{liu2025text2earth}, we adopt a progressive data refinement strategy: after pre-training on the full Git-10M corpus, we further refine GeoCore-9B on the high-quality subset whose quality scores exceed $4.8$~\cite{liu2025text2earth}, improving visual fidelity and fine-grained details. Training uses DeepSpeed ZeRO-2~\cite{rajbhandari2020zero} and takes approximately 15 days on eight NVIDIA Blackwell B200 GPUs.

\begin{figure}[t]
  \centering
  \includegraphics[width=1.0\textwidth]{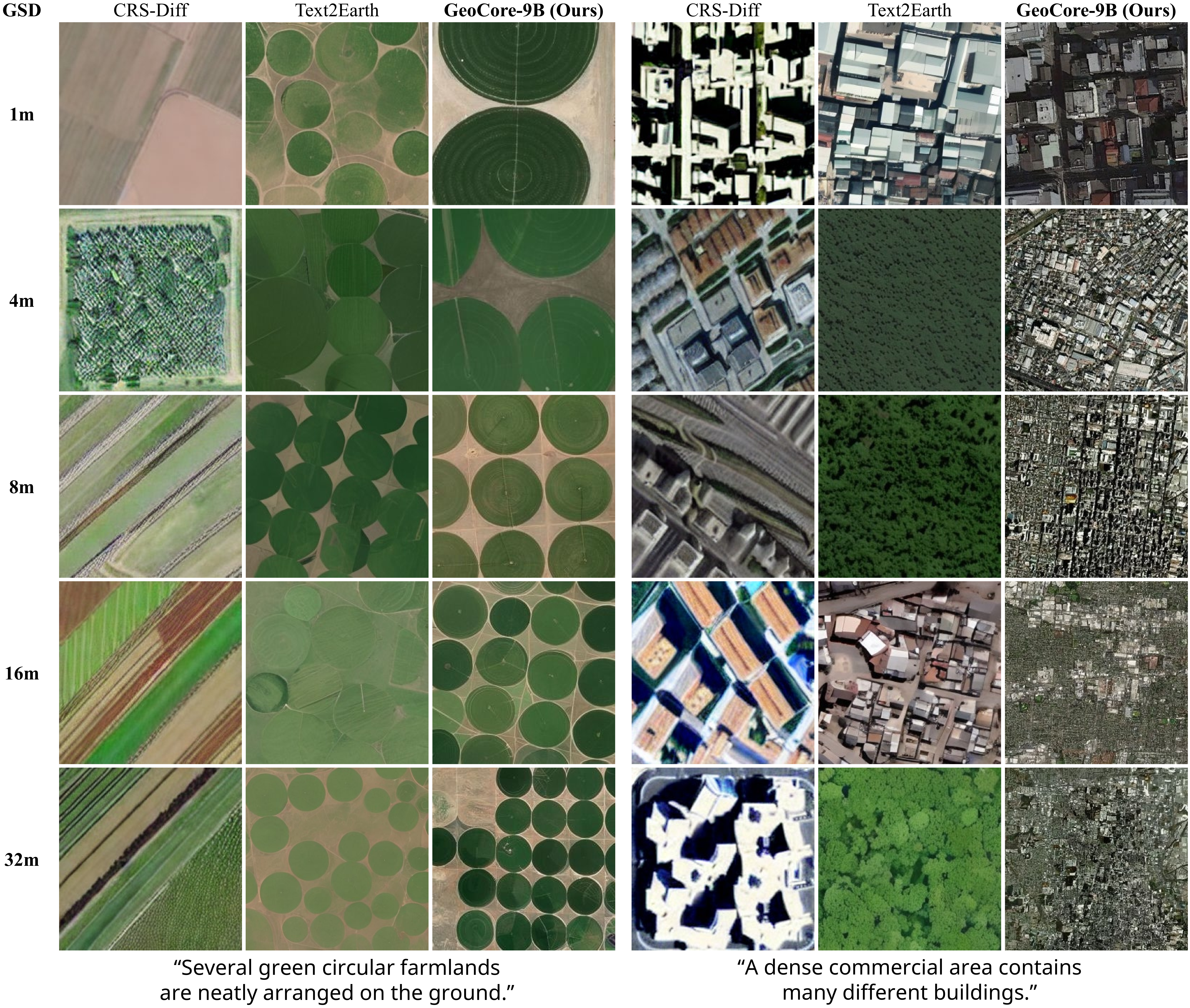}
    \caption{
    Qualitative comparison across varying ground sample distances (GSD). GeoCore-9B adaptively adjusts visual granularity from fine structural details (1$m$) to broad land-cover patterns (32$m$), demonstrating superior scale-awareness compared to CRS-Diff \cite{tang2024crs} and Text2Earth \cite{liu2025text2earth} which struggle with unnatural textures and scale inconsistency.
    }
    \label{fig:gsd_cond}
\end{figure}

\begin{figure}[t]
  \centering
  \includegraphics[width=1.0\textwidth]{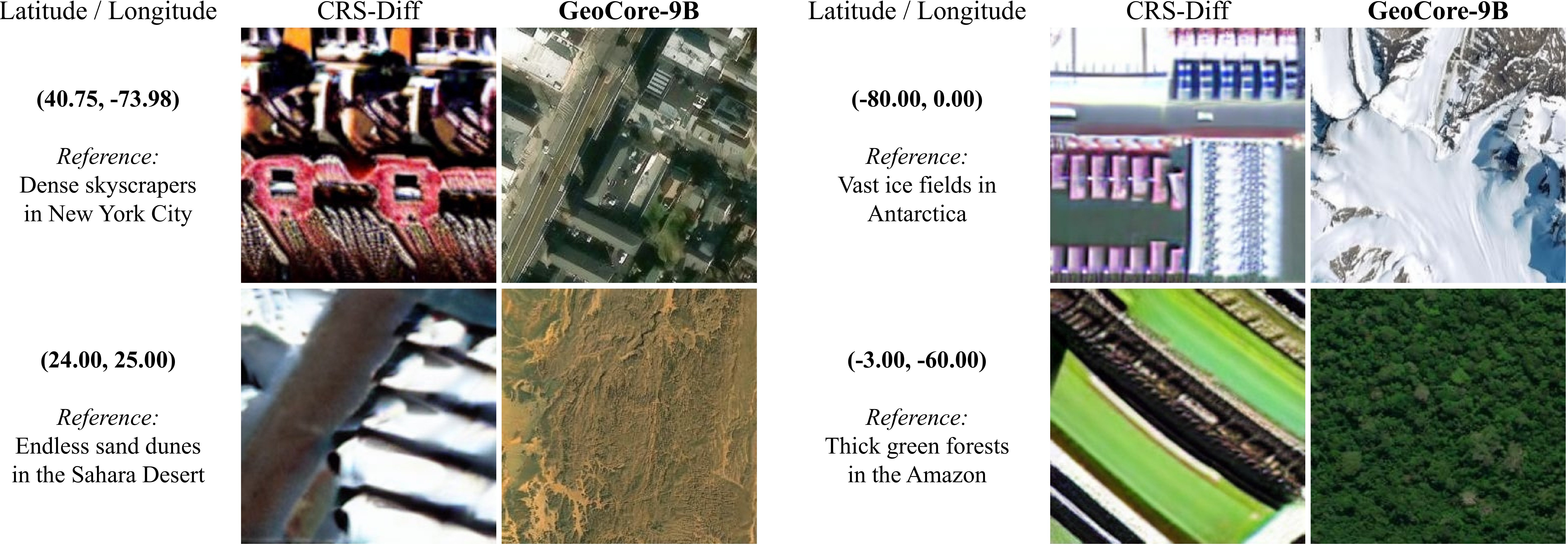}
    \caption{
    Qualitative comparison of text-free generation guided solely by latitude and longitude coordinates. Without text prompts, the baseline model (CRS-Diff \cite{tang2024crs}) fails to generate meaningful satellite imagery, suffering from severe artifacts and repeating patterns. In contrast, our proposed GeoCore-9B successfully retrieves geographic priors and synthesizes highly accurate terrains corresponding to the given coordinates.
    }
    \label{fig:coord_cond}
\end{figure}

\textbf{Downstream adaptation.}
For downstream tasks, we perform parameter-efficient adaptation by freezing the pre-trained GeoCore-9B backbone and optimizing lightweight LoRA adapters~\cite{hu2022lora}.
We use a rank of $r_{\mathrm{LoRA}}=64$ and a scaling factor of $\alpha_{\mathrm{LoRA}}=128$, injecting LoRA into the linear layers of the attention and feed-forward modules.
The adapters are optimized with a learning rate of $2\times10^{-4}$ and a batch size of 256.
For image-conditioned tasks, the condition images and target RGB images are encoded by the frozen VAE encoder, and the condition latent is concatenated with the noisy target latent $\mathbf{z}_t$ before the first input projection layer. We additionally optimize this input projection layer to accommodate the enlarged conditional latent input.

\textbf{Inference.}
We use a first-order Euler solver for the Flow Matching ODE~\cite{lipman2022flow, liu2022flow} with 50 sampling steps. Classifier-free guidance is applied with a scale of $w=4.0$, using an empty text prompt for text and learned null embeddings for geospatial metadata.

\subsection{Zero-shot Image Generation}

We first evaluate GeoCore-9B without task-specific fine-tuning to examine whether pre-training learns geo-aware generative priors. Given a text description and geospatial metadata, the model generates RGB images conditioned on semantic contents, physical scales, and geographic locations.

\textbf{Text.} To evaluate semantic controllability, we vary the text prompts while fixing the geospatial metadata. As shown in Fig.~\ref{fig:text_cond}, GeoCore-9B accurately follows prompts describing diverse scenes (e.g., residential areas, coastal regions, and industrial zones) while maintaining the authentic orthographic structures of satellite imagery. In contrast, baseline models such as CRS-Diff \cite{tang2024crs} and Text2Earth \cite{liu2025text2earth} often suffer from severe structural artifacts or unnatural textures.

\textbf{GSD.} To assess scale controllability, we vary the GSD values while fixing the text prompts and geographic coordinates. As illustrated in Fig.~\ref{fig:gsd_cond}, GeoCore-9B adaptively adjusts visual granularity according to the physical resolution. It produces finer structural details at lower GSD values (e.g., 1m) and coarser, broader land-cover patterns at higher GSD values (e.g., 32m), demonstrating superior scale-awareness compared to the baselines \cite{tang2024crs, liu2025text2earth}.

\textbf{Geographic coordinates.} To rigorously evaluate geographic conditioning, we challenge the models with a strictly text-free input setting: latitude and longitude coordinates \textit{only}. As shown in Fig.~\ref{fig:coord_cond}, while the baseline model (CRS-Diff \cite{tang2024crs}) fails to generate meaningful imagery without text prompts, suffering from severe artifacts and repeating patterns, GeoCore-9B successfully retrieves location-dependent geographic priors. It synthesizes highly accurate terrains corresponding precisely to the given coordinates.

\begin{figure}[t]
  \centering
  \includegraphics[width=1.0\textwidth]{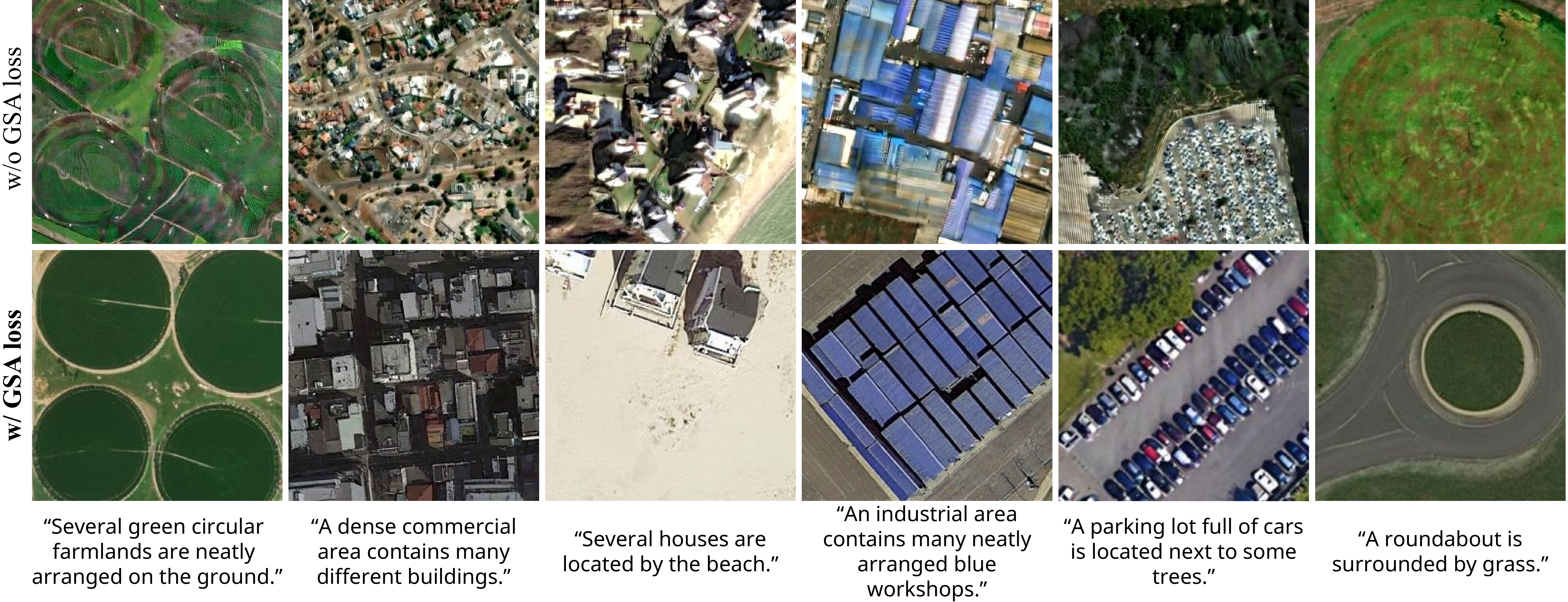}
    \caption{
    Ablation study on the Geospatial Semantic Alignment (GSA) loss. The inclusion of GSA loss (bottom row) improves structural fidelity and geometric consistency compared to the baseline without alignment (top row), which suffers from distorted boundaries and fragmented patterns.}
    \label{fig:ablation}
\end{figure}

\subsection{Ablation Study}
\label{subsec:ablation}

We evaluate the effect of Geospatial Semantic Alignment (GSA) by training a variant without the GSA objective, i.e., $\mu=0$.
As shown in Fig.~\ref{fig:ablation}, removing GSA leads to fragmented textures and distorted boundaries, especially for structured scenes such as circular farmlands, industrial roofs, parking lots, and roundabouts.
In contrast, GSA produces cleaner layouts and sharper object boundaries by aligning intermediate DiT features with satellite-specialist representations.
Fig.~\ref{fig:fid_ablation} further shows that GSA consistently reduces FID on a 10K Git-10M subset across training iterations, indicating faster convergence and improved structural fidelity without additional inference cost.

\subsection{RSICD Adaptation}
\label{subsec:rsicd_finetuning}

\begin{figure}[t]
  \centering
  % --- 왼쪽: 그림 (Ablation Study) ---
  \begin{minipage}{0.48\textwidth}
    \centering
    \includegraphics[width=\textwidth]{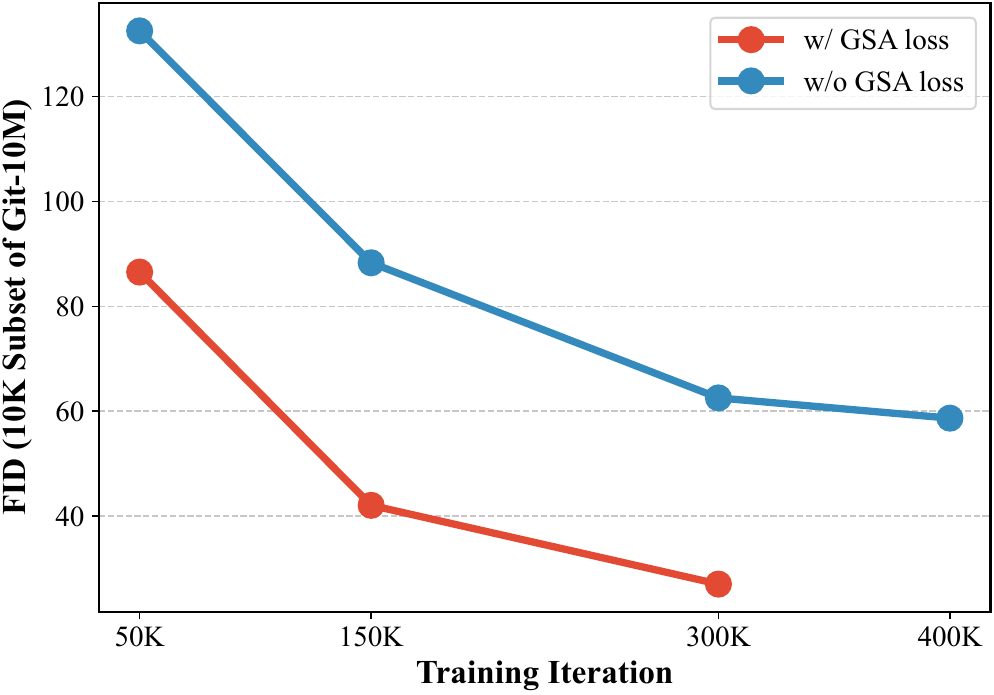}
    \caption{Ablation study on the Geospatial Semantic Alignment (GSA) loss. The inclusion of GSA loss improves structural fidelity.}
    \label{fig:fid_ablation}
  \end{minipage}
  \hfill
  % --- 오른쪽: 표 (Comparison Table) ---
  \begin{minipage}{0.48\textwidth}
    \centering
    \scriptsize
    \captionof{table}{Comparison with previous text-to-image methods on the RSICD \cite{lu2017exploring} dataset. \textbf{Bold} indicates the best result.}
    \label{tab:rsicd_comparison}
    \resizebox{\textwidth}{!}{%
    \begin{tabular}{lccc}
    \toprule
    \textbf{Method} & \textbf{IS $\uparrow$} & \textbf{FID $\downarrow$} & \textbf{CLIP $\uparrow$} \\
    \midrule
    Attn-GAN~\cite{xu2018attngan}        & 11.71 & 95.81  & 20.19 \\
    DAE-GAN~\cite{ruan2021dae}           & 7.71  & 93.15  & 19.69 \\
    StrucGAN~\cite{zhao2021text}         & 5.84  & --     & --    \\
    DF-GAN~\cite{tao2022df}              & 9.51  & 109.41 & 19.76 \\
    Lafite~\cite{zhou2022towards}        & 10.70 & 74.11  & 22.52 \\
    DALL-E~\cite{ramesh2021zero}         & 2.59  & 191.93 & 20.13 \\
    Txt2Img-MHN~\cite{xu2023txt2img}     & 5.99  & 102.44 & 20.27 \\
    \midrule
    RSDiff~\cite{sebaq2024rsdiff}        & 7.22  & 66.49  & --    \\
    CRS-Diff~\cite{tang2024crs}          & 18.39 & 50.72  & 20.33 \\
    Text2Earth~\cite{liu2025text2earth}  & --    & 24.49  & 25.62 \\
    \rowcolor{orange!12}
    \textbf{GeoCore-9B (Ours)}           & \textbf{22.15} & \textbf{18.82} & \textbf{27.15} \\
    \bottomrule
    \end{tabular}}
  \end{minipage}
\end{figure}

We evaluate the lightweight text-to-image adaptation capabilities of our model on the RSICD~\cite{lu2017exploring} dataset. Following prior works \cite{tang2024crs, liu2025text2earth}, we fine-tune LoRA adapters on the RSICD image-text training pairs. Since RSICD does not provide GSD values, latitudes, or longitudes, we employ learned null geospatial embeddings during both fine-tuning and inference. As shown in Table~\ref{tab:rsicd_comparison}, GeoCore-9B significantly outperforms previous text-to-image methods on the RSICD benchmark across all evaluated metrics (Inception score, FID score, and CLIP score).

\begin{table}[tbp]
    \scriptsize
    \centering
    \caption{
    Practical downstream adaptation results of GeoCore-9B. 
    We evaluate cloud removal on Sen2-MTC~\cite{huang2022ctgan} and SAR-to-optical cross-modal translation on QXS-SAROPT~\cite{huang2021qxs}.
    GeoCore-9B is adapted by simple fine-tuning and compared with task-specific or adapted baselines.
    \textbf{Bold} and \underline{underline} indicate the best and second-best results, respectively.
    }
    \label{tab:practical_downstream_adaptation}
    \vspace{0.2cm}
    \begin{minipage}[t]{0.46\textwidth}
        \centering
        {\small (a) Cloud Removal}\\[2pt]
        \resizebox{\linewidth}{!}{%
        \def\arraystretch{1.0}
        \setlength{\tabcolsep}{4pt}
        \begin{tabular}{lccc}
        \toprule
        \textbf{Methods} 
        & \textbf{PSNR$\uparrow$} 
        & \textbf{SSIM$\uparrow$} 
        & \textbf{LPIPS$\downarrow$} \\
        \midrule
        \multicolumn{4}{l}{\textit{\textbf{Task-specific specialist} methods}}\\[1pt]
        McGAN~\cite{enomoto2017filmy}              & 17.448 & 0.513 & 0.447 \\
        Pix2Pix~\cite{isola2017image}              & 16.985 & 0.455 & 0.535 \\
        DSen2-CR~\cite{meraner2020cloud}           & 16.827 & 0.534 & 0.446 \\
        STGAN~\cite{sarukkai2020cloud}             & 18.152 & 0.587 & 0.513 \\
        CTGAN~\cite{huang2022ctgan}                & 18.308 & 0.609 & 0.384 \\
        CR-TS-Net~\cite{ebel2022sen12ms}           & 18.585 & 0.615 & 0.342 \\
        PMAA~\cite{zou2023pmaa}                    & 18.369 & 0.614 & 0.392 \\
        UnCRtainTS~\cite{ebel2023uncrtaints}       & 18.770 & 0.631 & 0.333 \\
        DDPM-CR~\cite{jing2023denoising}           & 18.742 & 0.614 & 0.329 \\
        DiffCR~\cite{zou2024diffcr}                & 19.150 & 0.671 & 0.291 \\
        EMRDM~\cite{liu2025effective}              & \underline{20.067} & \underline{0.709} & \textbf{0.255} \\
        \midrule
        \rowcolor{orange!12}
        \multicolumn{4}{l}{\textit{\textbf{Foundation model adaptation}}}\\[1pt]
        \rowcolor{orange!12}
        \textbf{GeoCore-9B (Ours)}                 & \textbf{20.809} & \textbf{0.799} & \underline{0.256} \\
        \bottomrule
        \end{tabular}}
    \end{minipage}
    \hfill
    \begin{minipage}[t]{0.52\textwidth}
        \centering
        {\small (b) SAR-to-Optical Image Translation}\\[1pt]
        \resizebox{\linewidth}{!}{%
        \def\arraystretch{1.0}
        \setlength{\tabcolsep}{4pt}
        \begin{tabular}{lcccc}
        \toprule
        \textbf{Methods} 
        & \textbf{FID$\downarrow$} 
        & \textbf{LPIPS$\downarrow$} 
        & \textbf{HF-SCC$\uparrow$} 
        & \textbf{SSIM$\uparrow$} \\
        \midrule
        \multicolumn{5}{l}{\textit{\textbf{Task-specific or adapted} baselines}}\\[1pt]
        Pix2Pix~\cite{isola2017image}              & 196.89 & 0.454 & 0.0000 & 0.247 \\
        CycleGAN~\cite{zhu2017unpaired}            & 195.38 & 0.455 & 0.0001 & 0.251 \\
        SAR-SMTNet~\cite{youk2023transformer}      & 117.69 & 0.435 & 0.0003 & 0.260 \\
        CFCA-SET~\cite{lee2023cfca}                & 79.06  & 0.406 & 0.0006 & 0.273 \\
        BBDM~\cite{li2023bbdm}                     & 65.15  & 0.522 & 0.0004 & 0.238 \\
        ControlNet~\cite{zhang2023adding}          & 22.39  & 0.434 & 0.0001 & 0.257 \\
        Uni-ControlNet~\cite{zhao2023uni}          & 22.48  & 0.437 & 0.0002 & 0.257 \\
        StegoGAN~\cite{wu2024stegogan}             & 85.60  & 0.391 & 0.0019 & 0.280 \\
        DGDM~\cite{yoon2023deterministic}          & 147.23 & 0.634 & 0.0001 & 0.288 \\
        cBBDM~\cite{kim2025conditional}            & 69.47  & 0.420 & 0.0023 & 0.304 \\
        C-DiffSET~\cite{do2024c}                   & \underline{18.15} & \textbf{0.293} & \underline{0.0108} & \textbf{0.372} \\
        \midrule
        \rowcolor{orange!12}
        \multicolumn{5}{l}{\textit{\textbf{Foundation model adaptation}}}\\[1pt]
        \rowcolor{orange!12}
        \textbf{GeoCore-9B (Ours)}                 & \textbf{12.05} & \underline{0.377} & \textbf{0.3360} & \underline{0.370} \\
        \bottomrule
        \end{tabular}}
    \end{minipage}
\end{table}

\subsection{Practical and Challenging Downstream Tasks}
\label{subsec:practical_downstream}

Beyond text-to-image generation and controllability-oriented proxy tasks, we evaluate whether GeoCore-9B can be adapted to practical but challenging applications.
We consider two image-conditioned tasks: cloud removal and SAR-to-optical image translation.
For both tasks, GeoCore-9B is fine-tuned with LoRA while the pre-trained backbone remains frozen.

\begin{figure}[t]
  \centering
  \includegraphics[width=0.8\textwidth]{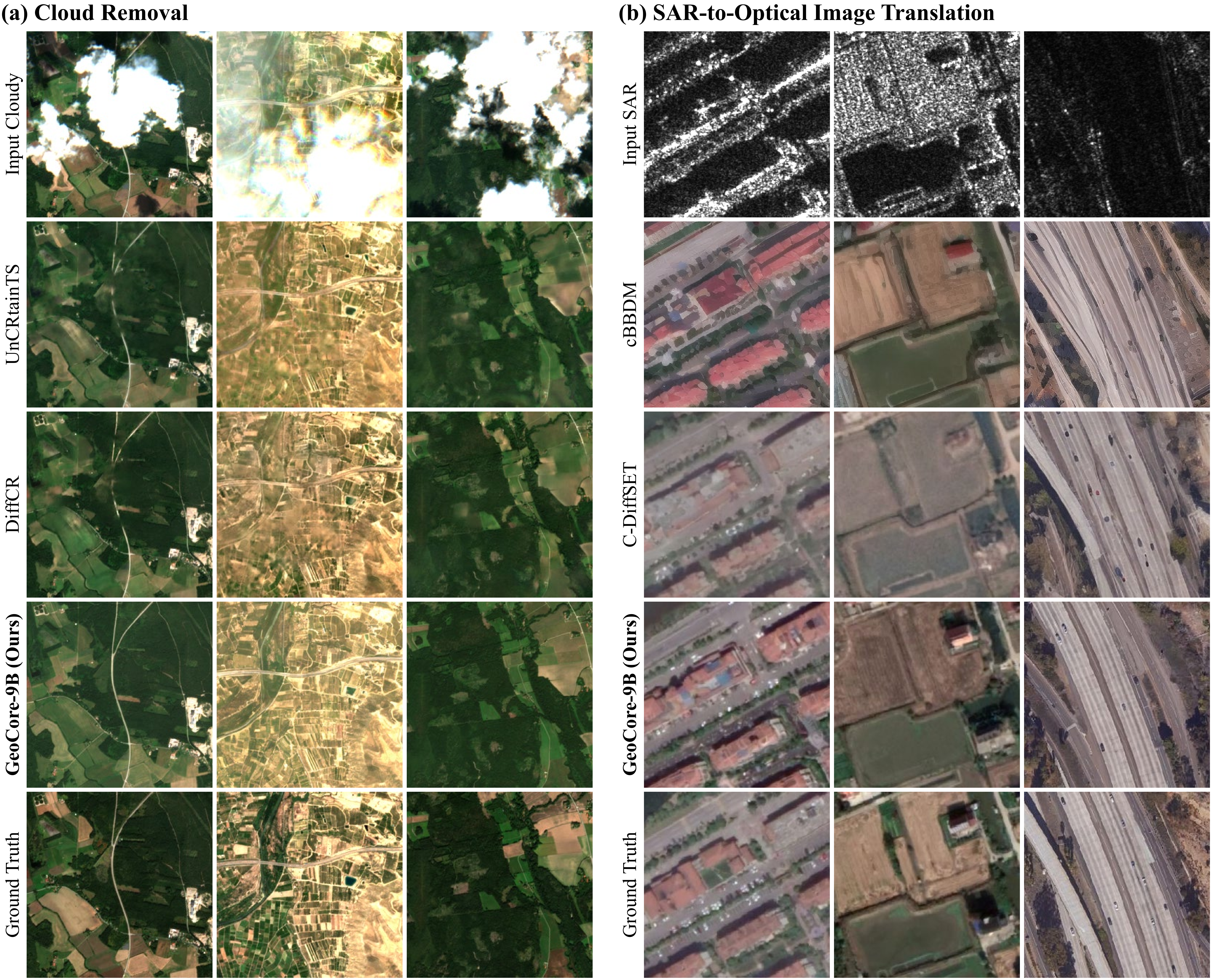}
    \caption{
      Qualitative comparison on practical downstream tasks.
      (a) Cloud removal: GeoCore-9B effectively removes heavy cloud contamination and reconstructs underlying structures (e.g., roads and field boundaries) much more faithfully than specialist baselines (UnCRtainTS \cite{ebel2023uncrtaints} and DiffCR \cite{zou2024diffcr}). 
      (b) SAR-to-optical image translation: GeoCore-9B translates noisy SAR inputs into realistic optical images, producing sharper man-made structures and more accurate spatial layouts compared to recent translation models (cBBDM \cite{kim2025conditional} and C-DiffSET \cite{do2024c}).
      }
    \label{fig:qual_cloud_removal}
\end{figure}

\textbf{Cloud removal.}
We evaluate GeoCore-9B on Sen2-MTC~\cite{huang2022ctgan}, where the model reconstructs cloud-free RGB images from cloudy observations (satellite RGB input images).
Table~\ref{tab:practical_downstream_adaptation}(a) shows quantitative comparisons of GeoCore-9B with task-specific specialist methods for the cloud removal task. GeoCore-9B outperforms task-specific cloud removal methods in PSNR and SSIM, while achieving LPIPS comparable to the strongest specialist baseline \cite{zou2024diffcr, liu2025effective}.
Fig.~\ref{fig:qual_cloud_removal}(a) shows that GeoCore-9B removes cloud contamination while preserving roads, and field boundaries.

\textbf{SAR-to-optical image translation.}
We further evaluate GeoCore-9B on QXS-SAROPT~\cite{huang2021qxs} for the SAR-to-optical translation task in Table~\ref{tab:practical_downstream_adaptation}(b).
GeoCore-9B achieves the best FID and HF-SCC, and remains comparable to the task-specific C-DiffSET baseline \cite{do2024c} in SSIM.
Fig.~\ref{fig:qual_cloud_removal}(b) shows that GeoCore-9B generates EO images with clearer man-made structures and more faithful spatial layouts than recent translation baselines.
These results demonstrate that GeoCore-9B can compete with, or outperform, task-specific models on such practical restoration and cross-modal translation tasks.

\subsection{Limitations and discussions}
\label{subsec:limitations}

GeoCore-9B leaves several directions for further extension.
First, although we validate its transferability on practical tasks such as cloud removal and SAR-to-optical translation, broader evaluations remains as future work for additional real-world applications such as pan-sharpening, super-resolution, and segmentation-conditioned generation.
Second, GeoCore-9B adopts a pre-trained VAE for efficient latent-space training for which future work will explore specialized latent learning for satellite imagery by jointly optimizing the VAE encoder with the DiT backbone, motivated by recent advances in representation and latent-space alignment~\cite{leng2025repa, yao2025reconstruction}.
Finally, GeoCore-9B can be extended to handle multispectral satellite imagery based upon the large datasets with geo metadata and rich text prompts.

\section{Conclusion}

We presented GeoCore-9B, a 9-billion-parameters generative foundation model trained from the scratch for Earth Observation.
Built upon a Flow Matching-based Diffusion Transformer, GeoCore-9B conditions generation on text and geospatial metadata, reducing reliance on natural image priors.
To stabilize training at this large scale, we introduced a Geospatial Semantic Alignment loss, which distills structural Earth surface priors from a frozen DINOv3-Sat teacher network during training without adding inference overhead.
Experiments show that GeoCore-9B achieves strong geo-aware generation capability and can be efficiently adapted to practical downstream tasks, including cloud removal and SAR-to-optical translation.
These results suggest that large-scale generative pre-training on satellite RGB data provides a promising foundation for geo-aware remote sensing generation.

\section*{Acknowledgments}

This work was supported by the National Research Foundation of Korea (NRF)
grant funded by the Korean government (MSIT) under the Sejong Science
Fellowship Program (RS-2026-25484549), for the project ``Visualizing the
Invisible Earth: A Reliability-Aware All-in-One SAR Analysis Framework with
Foundation Models.''

%%%%%%%%%%%%%%%%%%%%%%%%%%%%%%%%%%%%%%%%%%%%%%%%%%%%%%%%%%%%
\clearpage
\appendix

\section{Broader Impacts}
\label{app:broader_impacts}

The development of GeoCore-9B presents both significant positive potential and dual-use risks. On the positive side, our generative foundation model can substantially advance Earth Observation (EO) applications by providing high-quality data augmentation for scarce regions, and by enhancing downstream tasks such as environmental monitoring, disaster response, and urban planning. Conversely, the ability to generate highly realistic, geo-aware synthetic satellite imagery introduces the risk of creating geographical deepfakes. If misused, such technology could be exploited to generate disinformation regarding geopolitical events, natural disasters, or environmental conditions. To mitigate these negative societal impacts, we emphasize the necessity of developing robust synthetic image detection frameworks tailored specifically for satellite imagery and advocate for the responsible deployment and disclosure of generative EO models.

\section{Validation of the Pre-trained VAE on EO Data}
\label{app:vae_reconstruction}

GeoCore-9B operates in the latent space of a frozen variational autoencoder
(VAE)~\cite{labs2025flux} originally optimized for natural images.  The VAE
denotes the complete encoder--decoder model; below, $\mathcal{E}$ and
$\mathcal{D}$ denote its encoder and decoder, respectively.  For an input
$\mathbf{x}$, the reconstruction
$\widehat{\mathbf{x}}=\mathcal{D}(\mathcal{E}(\mathbf{x}))$ sets an upper bound
on input-faithful detail: structures discarded by $\mathcal{E}$ cannot be
reliably recovered by the DiT.  We therefore audit this bottleneck rather than
assuming that a natural-image VAE is lossless on EO imagery.

Under the submitted frozen-VAE encode--decode protocol, a random set of
100,000 Git-10M~\cite{liu2025text2earth} images gives 31.48 dB PSNR, 0.9310
SSIM, and 0.710 high-pass spatial correlation coefficient (HF-SCC).  These
averages indicate strong reconstruction fidelity and substantial preservation
of high-frequency structure at $256\times256$ resolution.  They do not,
however, separately test the most extreme frequency bands or guarantee the
preservation of small, sparse targets.

The positive HF-SCC values, ranging from 0.416 to 0.782 on the downstream
domains, support the VAE's practical use for the evaluated $256\times256$
tasks, including replicated-SAR inputs.  The lower Sen2-MTC values also expose
a genuine domain-dependent limitation.  After task-specific radiometric
preprocessing, many Sen2-MTC pixels and local variations have low contrast;
the natural-image VAE preserves the dominant coarse content but attenuates
some weak local variations.  This behavior is consistent with the lower
HF-SCC, although that aggregate metric alone cannot establish the precise
cause.  We therefore claim task- and resolution-bounded suitability, not
losslessness or guaranteed preservation of every EO microstructure.

\section{Additional Controlled Analyses}
\label{app:additional_analyses}

This section reports the controlled experiments completed after the original
submission.  We separate three questions: whether GSA helps at fixed scale,
whether the fixed checkpoint responds to metadata interventions, and whether
coordinate-only generations retrieve near-duplicates from the pre-training
corpus.  None of these experiments is presented as a full decomposition of
model scale, data scale, and compute.

\begin{figure}[t]
  \centering
  \includegraphics[width=\textwidth]{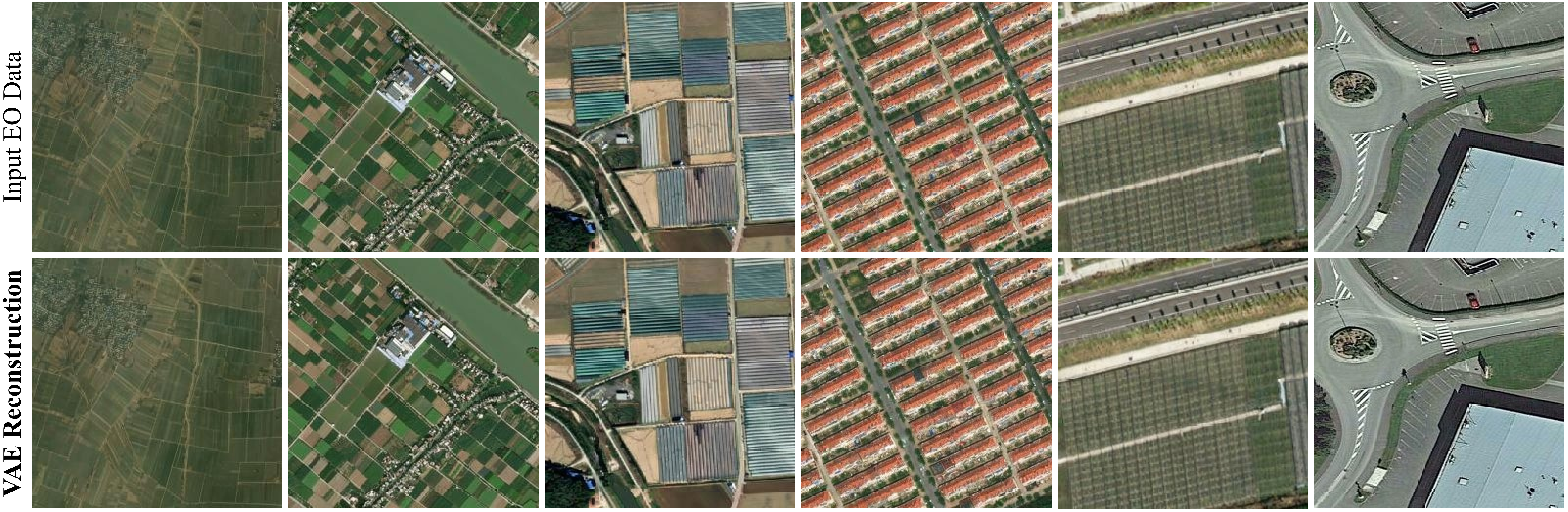}
    \caption{
    Qualitative evaluation of VAE reconstruction on Earth Observation data. Top row: Original input satellite images from Git-10M. Bottom row: Reconstructed images using the frozen pre-trained VAE. Despite the domain shift from natural images, the VAE accurately recovers fine-grained textures, complex building structures, and intricate field patterns without any EO-specific fine-tuning.
    }
    \label{fig:vae_recon}
\end{figure}

\subsection{Matched 9B Ablation of GSA}
\label{app:gsa_matched}

We compare two DiT backbones trained from random initialization with the same
9B architecture, Git-10M data, $256\times256$ resolution, pre-training budget,
and optimization protocol.  Their downstream adaptation schedules are also
identical; the controlled variable is only the GSA weight, $\mu=0$ versus
$\mu=0.5$.  The GSA teacher and projection head are discarded after
pre-training and add no inference-time module or cost.

\begin{table}[t]
  \centering
  \small
  \caption{Frozen-VAE encode--decode fidelity on the pre-training and
  downstream domains.  SAR intensities are replicated from one channel to
  three channels before VAE encoding.}
  \label{tab:vae_cross_domain}
  \vspace{0.2cm}
    \resizebox{0.8\textwidth}{!}{%
    \def\arraystretch{1.0}
    \setlength{\tabcolsep}{5pt}
  \begin{tabular}{lccccc}
    \toprule
    \textbf{Domain} & \textbf{$n$} & \textbf{PSNR $\uparrow$}
      & \textbf{SSIM $\uparrow$} & \textbf{LPIPS $\downarrow$}
      & \textbf{HF-SCC $\uparrow$} \\
    \midrule
    Git-10M RGB & 100,000 & 31.48 & 0.931 & -- & 0.710 \\
    QXS-SAROPT optical & 2,000 & 38.93 & 0.968 & 0.011 & 0.766 \\
    QXS-SAROPT SAR (1ch $\rightarrow$ 3ch) & 2,000 & 28.77 & 0.932 & 0.023 & 0.782 \\
    Sen2-MTC cloudy & 687 & 35.48 & 0.943 & 0.013 & 0.416 \\
    Sen2-MTC cloud-free & 687 & 35.58 & 0.924 & 0.016 & 0.564 \\
    \bottomrule
  \end{tabular}}
\end{table}

\begin{table}[t]
  \centering
  \scriptsize
  \caption{Matched downstream comparison with and without GSA.  All settings
  other than the pre-training GSA weight are held fixed.  HF-SCC uses the
  corrected, baseline-consistent definition.}
  \vspace{0.2cm}
  \label{tab:gsa_downstream_ablation}
  \resizebox{1.0\textwidth}{!}{%
      \def\arraystretch{1.0}
    \setlength{\tabcolsep}{4pt}
  \begin{tabular}{lll}
    \toprule
    \textbf{Task} & \textbf{w/ GSA ($\mu=0.5$)} & \textbf{w/o GSA ($\mu=0$)} \\
    \midrule
    RSICD text-to-image
      & \textbf{22.15 IS} / \textbf{18.82 FID} / \textbf{27.15 CLIP}
      & 19.16 IS / 28.43 FID / 24.21 CLIP \\
    QXS-SAROPT translation
      & \textbf{12.05 FID} / \textbf{0.377 LPIPS} / \textbf{0.0163 HF-SCC} / \textbf{0.370 SSIM}
      & 19.92 FID / 0.436 LPIPS / 0.0098 HF-SCC / 0.324 SSIM \\
    Sen2-MTC cloud removal
      & \textbf{20.809 PSNR} / \textbf{0.799 SSIM} / \textbf{0.256 LPIPS}
      & 19.553 PSNR / 0.683 SSIM / 0.284 LPIPS \\
    \bottomrule
  \end{tabular}}
\end{table}

GSA improves every reported metric.  In particular, it reduces FID by 9.61
points on RSICD and 7.87 points on QXS-SAROPT, while improving Sen2-MTC PSNR by
1.256 dB and SSIM by 0.116.  This matched comparison isolates a benefit from
GSA within the tested EO-trained 9B setting.  It does not isolate the effects
of overall model size, EO data, compute, or their interactions, and therefore
does not explain the entire margin to external baselines.

\subsection{Fixed-Checkpoint Metadata Interventions}
\label{app:metadata_intervention}

We conduct paired inference-time interventions on 1,000 metadata-parseable
Git-10M samples using the submitted zero-shot checkpoint.  For each sample, we
fix the caption, noise seed, checkpoint, Euler sampler, 50 sampling steps, and
CFG scale of 4.0.  We change only the metadata: full metadata, a learned-null
GSD, GSD shuffled from another sample, learned-null coordinates, or a jointly
shuffled latitude--longitude pair.

To measure whether the generated images reflect these interventions, we train
linear probes on frozen DINOv3-Sat ViT-L features from 20,000 real images and
validate on a disjoint set of 5,000 real images.  Before applying the probes to
generated images, the nine-bin GSD probe reaches 0.769 validation accuracy
(majority: 0.375), and the location probe reaches 0.585 over 81 eligible
$15^{\circ}$ regions (majority: 0.136).  Location results below use the 957
generated samples retained by the eligible-region criterion.

\begin{table}[t]
  \centering
  \small
  \caption{Paired metadata interventions.  ``Orig.'' and ``suppl.'' score a
  shuffled-condition output against its original and newly supplied metadata,
  respectively.  FID values support comparisons only within this protocol.}
  \label{tab:metadata_intervention}
    \vspace{0.2cm}
  \resizebox{0.8\textwidth}{!}{%
      \def\arraystretch{1.0}
    \setlength{\tabcolsep}{6pt}
  \begin{tabular}{lccc}
    \toprule
    \textbf{Condition} & \textbf{FID $\downarrow$}
      & \textbf{GSD-bin acc. $\uparrow$}
      & \textbf{$15^{\circ}$-region acc. $\uparrow$} \\
    \midrule
    Full metadata & 48.32 & 0.555 & 0.485 \\
    GSD null & 54.96 & 0.293 & 0.460 \\
    GSD shuffled & 49.46 & 0.286 (orig.) / 0.376 (suppl.) & 0.424 \\
    Coordinates null & 68.18 & 0.261 & 0.175 \\
    Coordinates shuffled & 51.20 & 0.492 & 0.110 (orig.) / 0.366 (suppl.) \\
    \bottomrule
  \end{tabular}}
\end{table}

Nulling GSD lowers GSD-bin accuracy from 0.555 to 0.293, and nulling coordinates
lowers region accuracy from 0.485 to 0.175.  The coordinate shuffle provides
the clearest intervention result: generated outputs agree more with the
supplied regions than with the original regions (0.366 versus 0.110).  For the
GSD shuffle, supplied-condition accuracy exceeds original-condition accuracy
(0.376 versus 0.286), but 0.376 is essentially the 0.375 majority baseline; we
therefore do not use this cell alone as evidence of fine-grained GSD following.
FID is numerically worse under every intervention, but these values are only
descriptive within-protocol checks.  Cross-field changes further indicate that
GSD and location are not perfectly disentangled.  Overall, the experiment
shows that metadata interventions affect the fixed model's outputs; it does
not quantify how much of the external-model performance margin arises from
metadata rather than scale, data, or GSA.

\subsection{Full-Corpus Near-Duplicate Retrieval}
\label{app:memorization_audit}

We encode all 10,503,567 pre-training images and 500 text-free,
coordinate-only generations with frozen DINOv3-Sat global features.  Before
examining the generated queries, we fix the similarity threshold at
$s_{\mathrm{thr}}=0.948$, the 95th percentile of nearest-\emph{other}-image
similarities from 1,000 real calibration queries with self-matches excluded
(calibration median: 0.882).  The generated-query nearest-neighbor
similarities have median 0.791, 95th percentile 0.871, and maximum 0.928;
none exceeds the threshold (0/500).  This finite global-feature test finds no
near-duplicate under the stated protocol, but it cannot exclude localized,
transformed, or other forms of memorization.  Moreover, coordinate-only
generation removes text but is not a complete text-null distributional
ablation.

\section{Exploratory Frozen-Feature Transfer Probes}
\label{app:frozen_probes}

Motivated by diffusion-feature probing in SatDiFuser~\cite{jia2025satdifuser},
we test whether task-relevant information is linearly accessible from the
submitted $256\times256$ GeoCore-9B checkpoint.  These experiments are
representation probes, not full task-specific systems.

We freeze the VAE and DiT.  To avoid conflict with the Flow Matching notation
in the main paper, let $\mathbf{z}_{\mathrm{clean}}=\mathcal{E}(\mathbf{x})$
and define the probe input at noise level $\tau$ as
\begin{equation}
  \mathbf{z}_{\tau}
  = (1-\tau)\mathbf{z}_{\mathrm{clean}} + \tau\boldsymbol{\epsilon},
  \qquad \boldsymbol{\epsilon}\sim\mathcal{N}(\mathbf{0},\mathbf{I}).
\end{equation}
We use empty text and null metadata, concatenate the $16\times16$ image-token
features from DiT blocks 4, 8, 16, and 24 into a 16,384-dimensional feature,
and train only one linear layer.  For dense tasks, the token grid is bilinearly
upsampled to $64\times64$.  We test the three fixed noise levels
$\tau\in\{0.25,0.50,0.75\}$ and use $\tau=0.50$ as the main reference because
it is the interpolation midpoint, not because it was tuned per task.

As a calibrated discriminative control, DINOv3-Sat ViT-L uses its standard
clean-input last-four-layer features.  The control follows the same images,
splits, token grid, upsampling, linear-head design, and training schedule,
although its feature width and extraction path differ from GeoCore-9B.

\begin{table}[t]
  \centering
  \small
  \caption{Frozen-feature linear probes on EuroSAT~\cite{helber2019eurosat},
  LoveDA~\cite{wang2021loveda}, and BRIGHT~\cite{chen2025bright}.  The
  train/validation sizes are shown in parentheses.}
  \vspace{0.2cm}
  \label{tab:frozen_feature_probes}
    \resizebox{0.8\textwidth}{!}{%
      \def\arraystretch{1.0}
  \setlength{\tabcolsep}{6pt}
  \begin{tabular}{llcccc}
    \toprule
    \textbf{Task (train/val)} & \textbf{Metric}
      & \textbf{$\tau=.25$} & \textbf{$\tau=.50$}
      & \textbf{$\tau=.75$} & \textbf{DINOv3-Sat} \\
    \midrule
    EuroSAT (12,960/3,240) & Top-1 & 97.3\% & 97.5\% & 97.3\% & \textbf{98.0\%} \\
    LoveDA (3,000/1,200) & mIoU & 0.328 & 0.382 & 0.282 & \textbf{0.451} \\
    BRIGHT (2,500/349 pairs) & mIoU & 0.402 & \textbf{0.482} & 0.381 & 0.442 \\
    \bottomrule
  \end{tabular}}
\end{table}

At $\tau=0.50$, GeoCore-9B is within 0.5 top-1 percentage points of the
DINOv3-Sat control on EuroSAT, reaches approximately 85\% of its LoveDA mIoU,
and exceeds it on BRIGHT (0.482 versus 0.442).  Repeating the BRIGHT linear
probe on the same frozen features gives 0.466 mIoU, still above the control.
Classification varies by only 0.2 points across the tested noise levels,
whereas both dense tasks perform best at the fixed midpoint.

These probes show linear accessibility of task-relevant information, but they
are not comparisons with full-resolution specialist systems.  DINOv3-Sat is
also the GSA teacher, and we have not run a matched w/o-GSA feature probe;
therefore, these results do not isolate how much of the transfer is caused by
GSA.  BRIGHT uses replicated-grayscale SAR only as an input, so this experiment
also does not demonstrate SAR generation.

\section{Implementation Details}

\textbf{3D RoPE.}
To effectively model the joint sequence of text and latent image tokens, we employ a 3D Rotary Positional Embedding (3D RoPE)~\cite{su2024roformer}. Specifically, we map each token into a unified 3D coordinate system $(x, y, z)$. For textual tokens, the $x$-axis represents the 1D sequence index $m$, while for visual tokens, the $(y, z)$ axes correspond to the 2D spatial grid coordinates $(i, j)$. This 3D formulation allows the model to inherently reason about relative distances both within and across modalities, maintaining robust geospatial structural reasoning even under dynamic variations in image resolution and text length.

\textbf{Classifier-Free Guidance.}
For Classifier-Free Guidance (CFG)~\cite{ho2022classifier}, we implement an independent condition dropout strategy during training. While the text prompt is simply replaced by an empty string, masking continuous geographic metadata requires a more robust formulation. Thus, we introduce explicit learnable null embeddings $e_{r}$, $e_{\phi}$, and $e_{\lambda}$ to substitute the missing GSD, latitude, and longitude features, respectively. With a probability $p_\text{cfg}$ (e.g., 0.1), all elements in the conditioning set $\mathcal{C} = \{d, r, \phi, \lambda\}$ are jointly replaced by their null counterparts to learn a fully unconditional prior. Otherwise, each condition is independently masked with its own dropout probability. This rigorous formulation ensures the model captures both the joint and marginal distributions of the geospatial and textual priors.

\section{More Results and Qualitative Diversity}

In this section, we provide additional qualitative results to further demonstrate the generative capabilities, diversity, and downstream transferability of GeoCore-9B.

\begin{figure}[t]
  \centering
  \includegraphics[width=\textwidth]{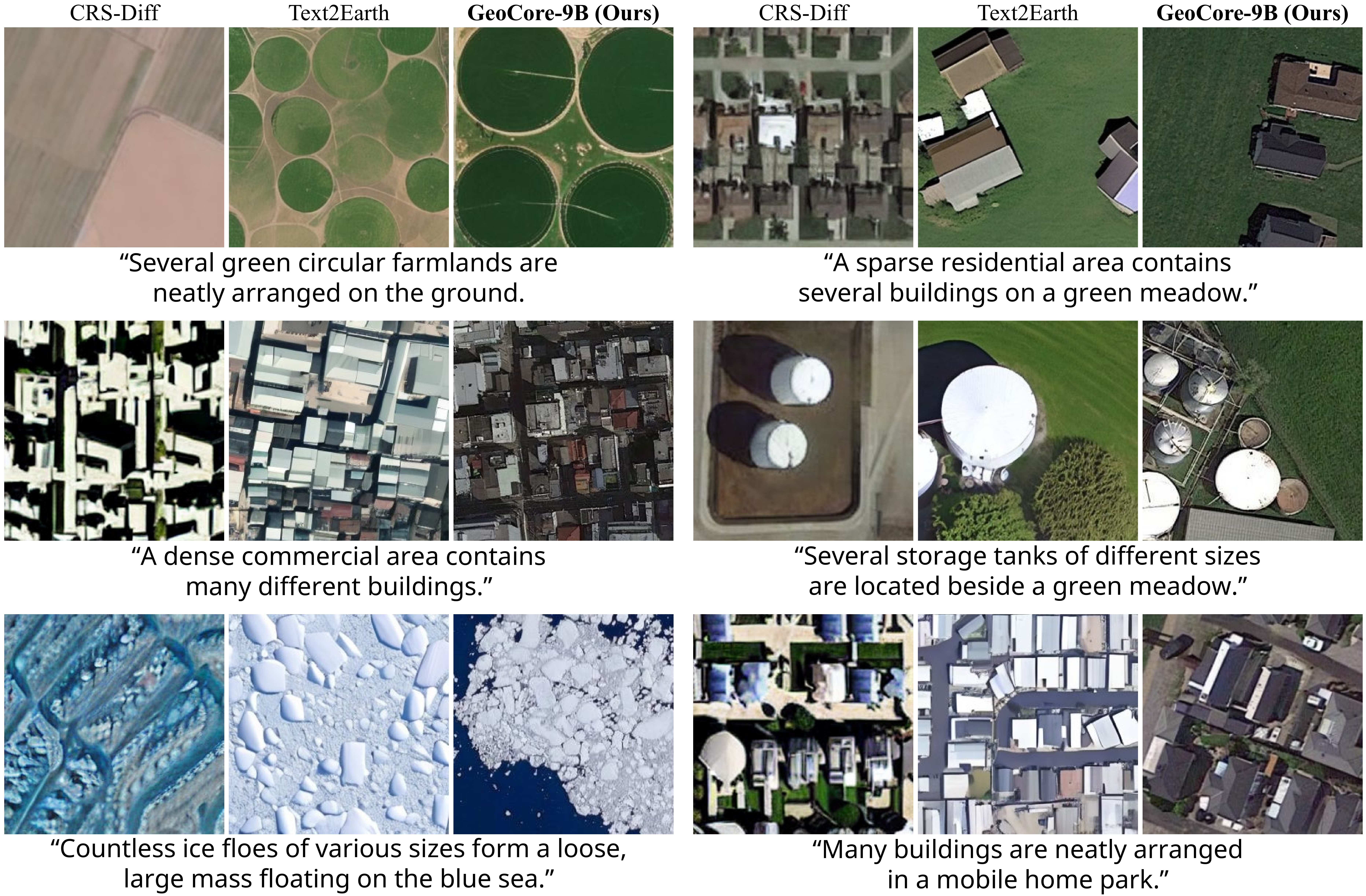}
\caption{
    Qualitative comparison of text-conditioned generation. Given various text descriptions, GeoCore-9B synthesizes highly realistic and structurally accurate satellite imagery, outperforming baselines (CRS-Diff \cite{tang2024crs} and Text2Earth \cite{liu2025text2earth}) which often suffer from severe artifacts.
    }
    \label{fig:text_2}
\end{figure}

\textbf{Text-conditioned generation.} Figure~\ref{fig:text_2} presents a broader set of text-conditioned generation examples. As shown, GeoCore-9B consistently synthesizes highly diverse and structurally realistic satellite imagery across various complex textual prompts. Compared to existing baseline models such as CRS-Diff \cite{tang2024crs} and Text2Earth \cite{liu2025text2earth}, which frequently exhibit unnatural textures and structural artifacts, our model strictly maintains the authentic orthographic geometry of Earth observation data without losing fine-grained details.

\begin{figure}[t]
  \centering
  \includegraphics[width=\textwidth]{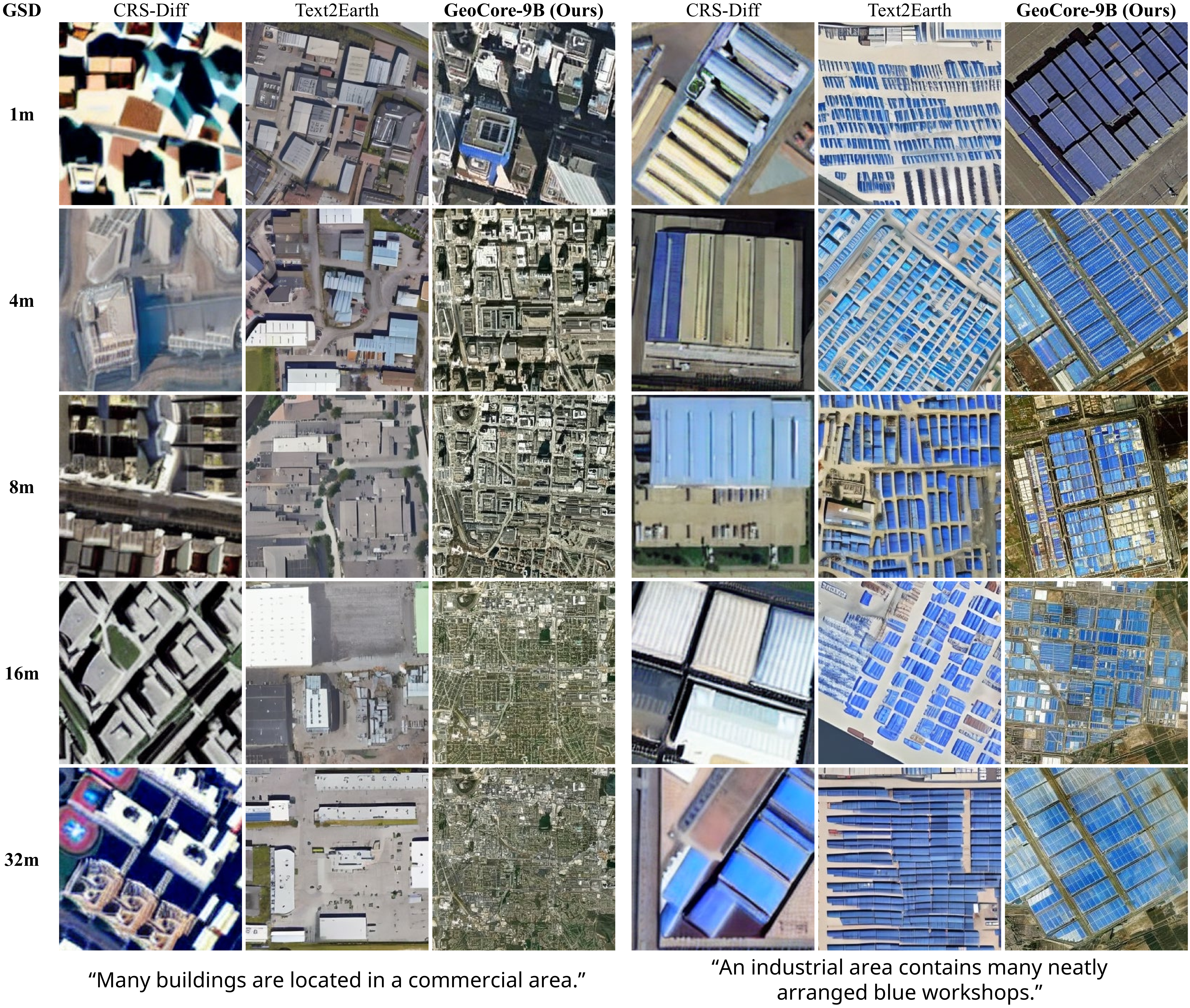}
    \caption{
    Qualitative comparison across varying ground sample distances (GSD). GeoCore-9B adaptively adjusts visual granularity from fine structural details (1$m$) to broad land-cover patterns (32$m$), demonstrating superior scale-awareness compared to CRS-Diff \cite{tang2024crs} and Text2Earth \cite{liu2025text2earth} which struggle with unnatural textures and scale inconsistency.
    }
    \label{fig:gsd_2}
\end{figure}

\begin{figure}[t]
  \centering
  \includegraphics[width=\textwidth]{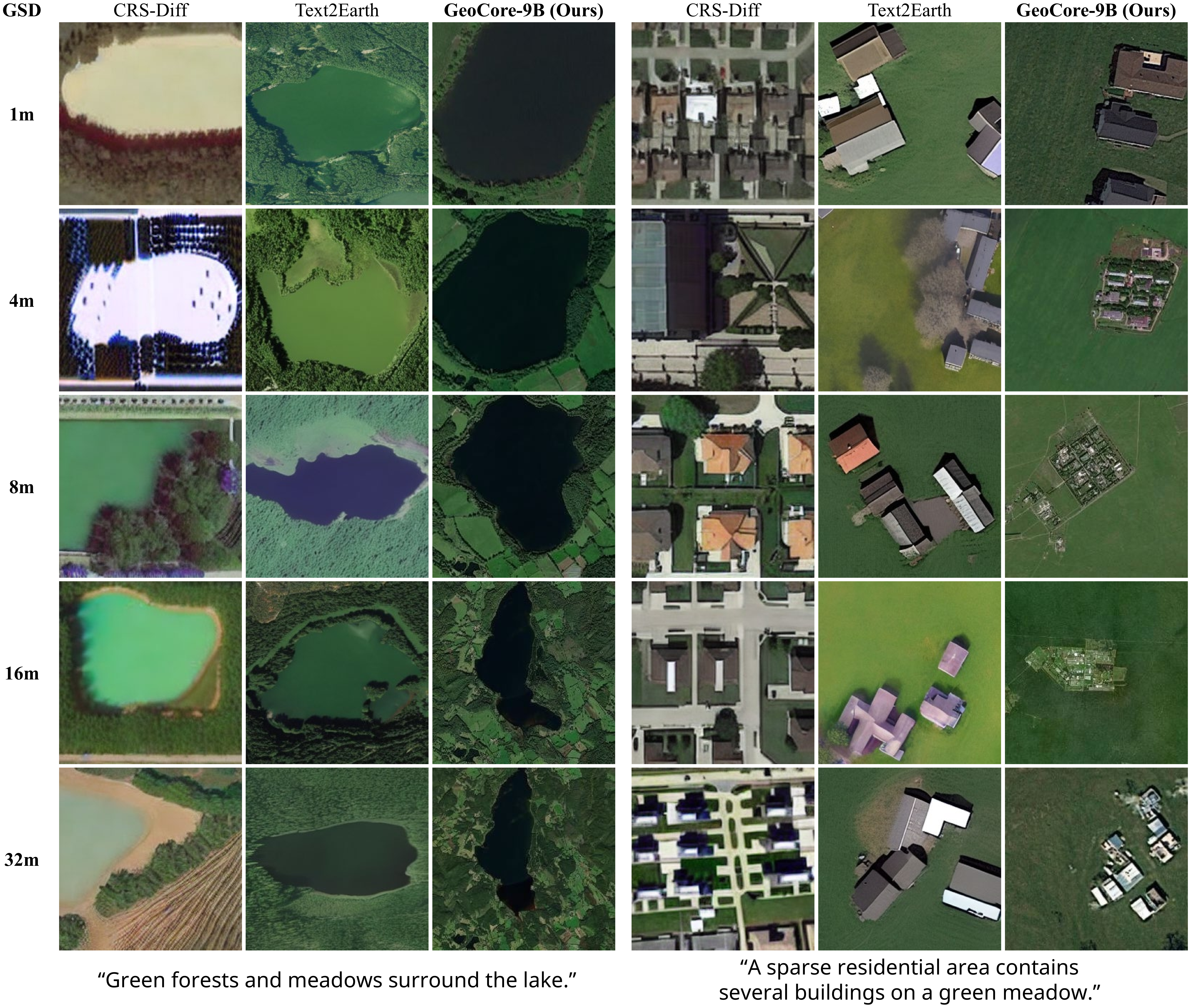}
    \caption{
    Qualitative comparison across varying ground sample distances (GSD). GeoCore-9B adaptively adjusts visual granularity from fine structural details (1$m$) to broad land-cover patterns (32$m$), demonstrating superior scale-awareness compared to CRS-Diff \cite{tang2024crs} and Text2Earth \cite{liu2025text2earth} which struggle with unnatural textures and scale inconsistency.
    }
    \label{fig:gsd_3}
\end{figure}

\begin{figure}[t]
  \centering
  \includegraphics[width=\textwidth]{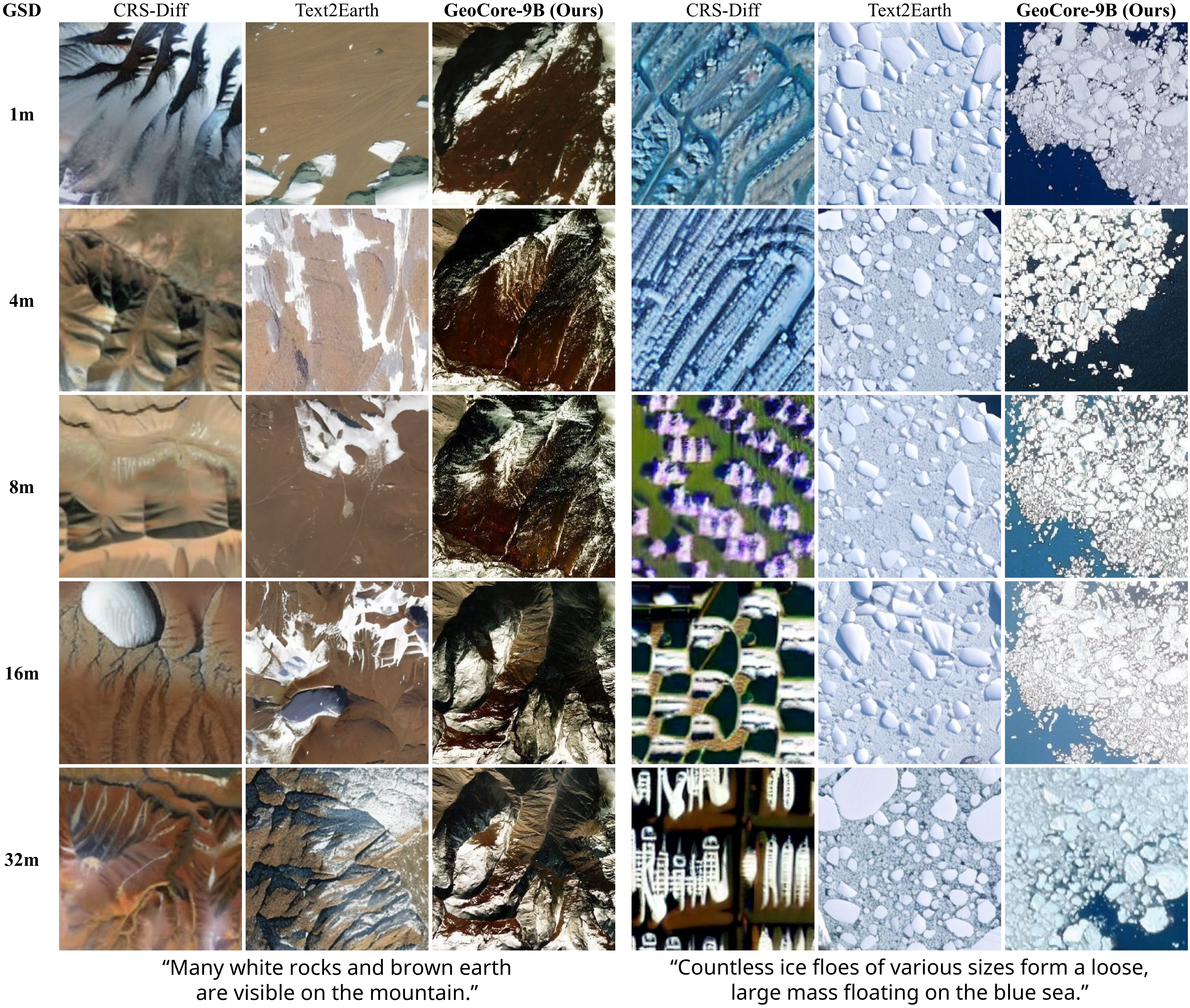}
    \caption{
    Qualitative comparison across varying ground sample distances (GSD). GeoCore-9B adaptively adjusts visual granularity from fine structural details (1$m$) to broad land-cover patterns (32$m$), demonstrating superior scale-awareness compared to CRS-Diff \cite{tang2024crs} and Text2Earth \cite{liu2025text2earth} which struggle with unnatural textures and scale inconsistency.
    }
    \label{fig:gsd_4}
\end{figure}

\textbf{GSD-conditioned generation.} Figure~\ref{fig:gsd_2}, Figure~\ref{fig:gsd_3}, and Figure~\ref{fig:gsd_4} provide further qualitative results demonstrating the scale-awareness of GeoCore-9B across varying ground sample distances (GSD). As the spatial resolution transitions from fine-grained (e.g., 1$m$) to coarse-grained (e.g., 32$m$), GeoCore-9B adaptively adjusts its visual granularity. It seamlessly shifts from synthesizing detailed individual objects to rendering broad, macroscopic land-cover patterns. Unlike baseline models that frequently struggle with scale inconsistency—producing unnaturally sized objects or repetitive textures at extreme resolutions—our model maintains strict physical and structural fidelity corresponding to the exact target GSD.

\begin{figure}[t]
  \centering
  \includegraphics[width=\textwidth]{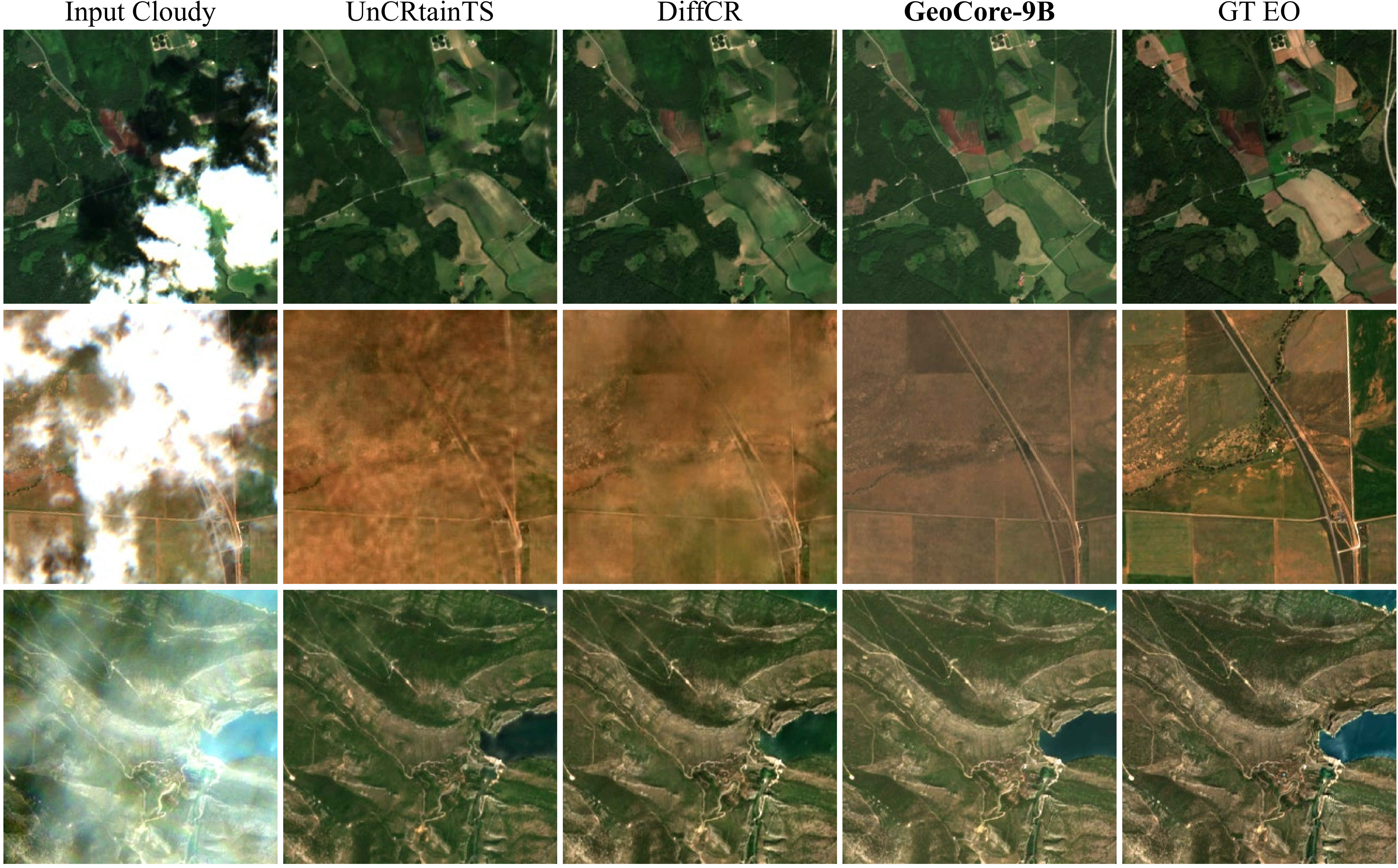}
    \caption{
      Qualitative comparison on practical downstream tasks (cloud removal).
      GeoCore-9B effectively removes heavy cloud contamination and reconstructs underlying structures (e.g., roads and field boundaries) much more faithfully than specialist baselines (UnCRtainTS \cite{ebel2023uncrtaints} and DiffCR \cite{zou2024diffcr}). 
      }
    \label{fig:cr_2}
\end{figure}

\textbf{Cloud removal.} Beyond zero-shot generation, we provide extended qualitative comparisons for practical downstream restoration tasks. Figure~\ref{fig:cr_2} illustrates additional results for the cloud removal task. Even under heavy and heterogeneous cloud coverage, GeoCore-9B accurately recovers the underlying geographic contexts, such as intricate road networks, detailed field boundaries, and diverse land-cover types. It notably produces much more faithful and sharper reconstructions than specialist models like UnCRtainTS \cite{ebel2023uncrtaints} and DiffCR \cite{zou2024diffcr}, which often yield blurry or semantically inconsistent regions.

\begin{figure}[t]
  \centering
  \includegraphics[width=\textwidth]{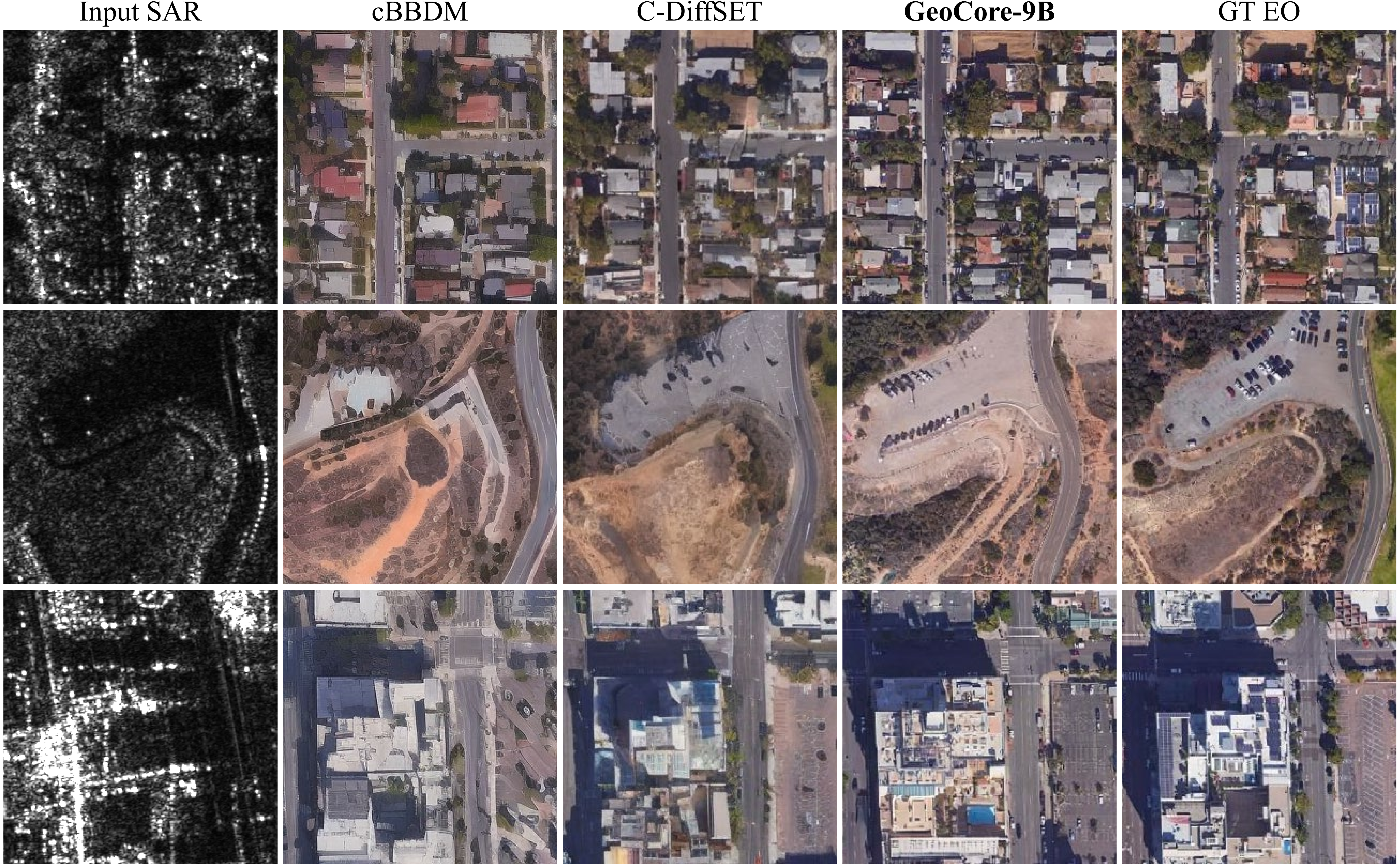}
    \caption{
      Qualitative comparison on practical downstream tasks (SAR-to-optical image translation). GeoCore-9B translates noisy SAR inputs into realistic optical images, producing sharper man-made structures and more accurate spatial layouts compared to recent translation models (cBBDM \cite{kim2025conditional} and C-DiffSET \cite{do2024c}).
      }
    \label{fig:sar_2}
\end{figure}

\textbf{SAR-to-optical image translation.} Finally, Figure~\ref{fig:sar_2} showcases more examples of SAR-to-optical cross-modal translation. The inherently noisy and speckle-heavy nature of Synthetic Aperture Radar (SAR) imagery makes this task particularly challenging. Nevertheless, GeoCore-9B effectively translates these noisy inputs into clear, high-fidelity optical images. It excels at generating accurate spatial layouts and sharp man-made structures, demonstrating clear visual superiority over recent state-of-the-art translation models, including cBBDM \cite{kim2025conditional} and C-DiffSET \cite{do2024c}.

\section{Scope, Attribution, and Limitations}
\label{app:scope_limitations}

\textbf{Contribution and attribution.}
GeoCore-9B uses a standard Flow Matching DiT backbone; we do not claim a new
generic Flow Matching objective or transformer block, and numerical
geospatial conditioning is not itself a new primitive.  The system
contribution is the EO-trained 9B generative backbone and its release.  The
specific design contributions are the integration of continuous GSD and
coordinate conditioning and GSA as an EO-specialist representation-alignment
instantiation.  The matched experiment in
Table~\ref{tab:gsa_downstream_ablation} isolates GSA at fixed 9B architecture,
data, and training scale, while the interventions in
Table~\ref{tab:metadata_intervention} establish fixed-model responsiveness to
metadata.  Neither experiment decomposes the effects of model size, EO data,
compute, metadata, and GSA relative to external systems.  A full factorial
study varying these factors at 9B scale remains outside the present scope.
Git-10M is a public dataset, and we do not claim its construction as a
contribution.

CRS-Diff~\cite{tang2024crs} is best interpreted as a controllable EO generator,
whereas Text2Earth~\cite{liu2025text2earth} is the more direct EO
text-to-image comparator.  Closed general-purpose generators do not expose
weights, training data, or fine-tuning access and therefore provide, at most,
uncontrolled qualitative references rather than matched quantitative
evidence.

\textbf{Residual natural-image components and the VAE ceiling.}
The 9B generative DiT backbone is initialized and trained from scratch on EO
data, but the complete system retains an off-the-shelf natural-image VAE and
pretrained text encoders.  Thus, the model avoids initialization of its DiT
from a natural-image diffusion backbone; it is not entirely free of
natural-image priors.  The VAE is also an information bottleneck whose
reconstruction quality upper-bounds input-faithful detail.  The audits in
Sec.~\ref{app:vae_reconstruction} support practical use for the current
$256\times256$ tasks but do not guarantee preservation of every tiny target or
high-frequency structure.  A promising direction is EO-specific end-to-end
adaptation of the VAE and DiT, following representation-aligned joint tuning
such as REPA-E~\cite{leng2025repa}, while retaining reconstruction and
EO-specific high-frequency preservation objectives.

\textbf{RGB output and sensor scope.}
GeoCore-9B currently generates RGB optical imagery.  In SAR-to-optical
translation, SAR is only an input condition and the target remains RGB
optical; the current model does not generate SAR or multispectral imagery.
DINOv3-Sat never directly processes SAR in the submitted system: GSA is used
only during RGB EO pre-training to supervise intermediate optical-target
representations, and the teacher and projection head are removed before
downstream adaptation and inference.  Consequently, SAR measurements are not
directly forced to match the RGB teacher space.  Nevertheless, GSA can leave
an optical prior in the learned backbone weights.  The improved QXS-SAROPT
results establish compatibility with the tested SAR-conditioned, RGB-output
setting, not sensor-universal transfer.  SARMAE~\cite{liu2026sarmae} provides
complementary evidence that optical DINOv3 supervision can benefit SAR
representation learning, but broader output modalities would still require
modality-appropriate latent encoders and modality-specific or multimodal
teachers.

\textbf{Discriminative transfer.}
The exploratory probes in Sec.~\ref{app:frozen_probes} show that task-relevant
information is linearly accessible from frozen GeoCore-9B features.  They use
a low-resolution token grid and a single linear head, rather than full
task-specific systems, and do not include a matched w/o-GSA probe.  They
therefore neither establish discriminative state of the art nor attribute the
observed transfer specifically to GSA.

%%%%%%%%%%%%%%%%%%%%%%%%%%%%%%%%%%%%%%%%%%%%%%%%%%%%%%%%%%%%

\clearpage
\bibliographystyle{plain}
\bibliography{neurips_2026}

\end{document}